\def\year{2022}\relax
\documentclass[letterpaper]{article} 
\usepackage{aaai22}  
\usepackage{times}  
\usepackage{helvet}  
\usepackage{courier}  
\usepackage[hyphens]{url}  
\usepackage{graphicx} 
\usepackage{natbib}  
\usepackage{caption} 
\DeclareCaptionStyle{ruled}{labelfont=normalfont,labelsep=colon,strut=off} 
\usepackage[utf8]{inputenc} 
\usepackage{booktabs}       
\usepackage{amsfonts}       
\usepackage{nicefrac}       
\usepackage{microtype}      
\usepackage{xcolor}         
\usepackage{subfigure}

\usepackage{amsthm}
\newtheorem{theorem}{Theorem}
\newtheorem{corollary}{Corollary}

\usepackage{amsmath}
\usepackage{mathtools}
\usepackage{amsfonts}  
\usepackage{algorithm}
\usepackage{algorithmic}
\usepackage{adjustbox}
\usepackage{caption}
\usepackage[normalem]{ulem}

\newcommand{\ep}{\varepsilon}
\renewcommand{\epsilon}{\varepsilon}
\definecolor{ed}{RGB}{225,0,0}

\usepackage{paralist}

\usepackage{bbm,dsfont}
\title{
iDECODe: In-distribution Equivariance for Conformal\\ Out-of-distribution Detection}

 \author{
 Ramneet Kaur\textsuperscript{\rm 1}, 
 Susmit Jha\textsuperscript{\rm 2}, 
 Anirban Roy\textsuperscript{\rm 2}, 
 Sangdon Park\textsuperscript{\rm 13}, 
 Edgar Dobriban\textsuperscript{\rm 4}, 
 Oleg Sokolsky\textsuperscript{\rm 1}, 
 Insup Lee\textsuperscript{\rm 1} 
 }
 \affiliations{
 \textsuperscript{\rm 1} Department of Computer and Information Science, University of Pennsylvania, Philadelphia, USA\\
 \textsuperscript{\rm 2} Computer Science Laboratory, SRI International, Menlo Park, USA\\
 \textsuperscript{\rm 3} School of Computer Science, Georgia Institute of Technology\\
 \textsuperscript{\rm 4} Statistics \& Computer Science, University of Pennsylvania\\
 \texttt{ramneetk@seas.upenn.edu,\{susmit.jha,anirban.roy\}@sri.com,} \\
 \texttt{dobriban@wharton.upenn.edu, \{sangdonp, sokolsky, lee\}@cis.upenn.edu}
 }

\begin{document}

\maketitle

\begin{abstract}
Machine learning methods such as deep neural networks (DNNs), despite their success across different domains, are known to often generate incorrect predictions with high confidence on inputs outside their training distribution. The deployment of DNNs in safety-critical domains requires detection of out-of-distribution (OOD) data so that DNNs can abstain from making predictions on those. A number of methods have been recently developed for OOD detection, but there is still room for improvement. We propose the new method iDECODe, leveraging in-distribution equivariance for conformal OOD detection. It relies on a novel base non-conformity measure and a new aggregation method, used in the inductive conformal anomaly detection framework, thereby guaranteeing a bounded false detection rate. We demonstrate the efficacy of iDECODe by experiments on image and audio datasets, obtaining state-of-the-art results. We also show that iDECODe can detect adversarial examples. 
\end{abstract}

\section{Introduction}
\label{sec:intro}
Powerful modern machine learning methods, such as deep neural networks (DNNs) exhibit remarkable performance in domains such as computer vision~\cite{img-classification}, audio recognition~\cite{speech-recog}, and natural language processing~\cite{text-analysis}. However, DNNs are known to generate overconfident and incorrect predictions on inputs outside their training distribution~\cite{baseline}. The responsible deployment of machine learning (ML) in safety-critical domains such as autonomous vehicles~\cite{ML-App7}, and medicine~\cite{NNclinically} requires detection of out-of-distribution (OOD) data, so that these ML models can abstain from making predictions on those. A great number of methods have been developed for OOD detection, but there is still significant room for improvement. In this paper, we propose iDECODe, a novel OOD detection method based on conformal prediction with transformation equivariance learned on in-distribution (iD) data.


Equivariance of outputs to certain geometric data transforms is a general desired property of ML systems. For example, it is desirable for a classifier trained on images of upright cats to also correctly classify rotated images of cats. In other words, classifiers should learn a representation that is invariant to the orientation of the training data. 
 Sharing of kernels in convolutional neural networks (CNNs), and more generally group CNNs, leads to learning features equivariant to translations, and more generally to group transforms. Therefore, group CNNs are mathematically guaranteed to be equivariant to translations for \emph{all inputs}, which has played a critical role in the success of CNNs~\cite{cnn_eq1, cnn_eq2}. Another common approach to encode these transformations is data augmentation~\cite{baird1992document,data_aug,chen2020group,chatzipantazis2021learning}. This is not guaranteed to lead to equivariance for all inputs, and is more likely to work for \emph{in-distribution data} used for training than for \emph{out-of-distribution data} dissimilar to that used for training. This is the crucial insight for us:
    we propose \emph{using deviations from equivariance to test OOD-ness}.

To get rigorous control on the false detection rate, we leverage conformal prediction~\cite{icp,cp}, which is a general methodology to test if an input conforms to the training data. It uses a non-conformity measure (NCM) to quantitatively estimate how different an input is from the training distribution. Commonly used NCMs are based on the properties of the input's $k$-nearest neighbors from the training data~\cite{cp, dknn} and kernel density estimation methods~\cite{kde}. Inductive conformal anomaly detection (ICAD)~\cite{icad} uses an NCM to assign a non-conformity score to the input for computing its $p$-value indicating anomalous behavior. The performance of ICAD can depend strongly on the choice of the NCM~\cite{cp}. We propose using the deviation (or error) in the predictable behavior of a model \textit{equivariant in-distribution (iD) with respect to a set $G$ of transformations} as the NCM for OOD detection. 


ICAD uses a single score from the NCM to compute the $p$-value. We instead propose using a vector of $n$ non-conformity scores computed from the proposed NCM with $n$ transformations sampled as independent and identically distributed (IID) variables from a distribution over $G$. Intuitively, with a single transformation, an OOD datapoint might behave as a transformed iD datapoint, but the likelihood of this decreases with the number of transformations $n$.

The contributions of this paper are summarized as follows:
\begin{compactitem}
    \item {\bf Novel base NCM.} We propose a novel base NCM for detecting the OOD nature of an input as the error in the iD equivariance learned by a model with respect to a set $G$ of transforms.  
    \item {\bf Novel aggregation method.} We propose a novel approach to increase performance by aggregating $n$ scores computed from the proposed base NCM on $n$ IID transformations sampled from a distribution over $G$, leading to an aggregated NCM. 
    \item {\bf iDECODe.} Using aggregated NCM in the ICAD framework leads to our proposed iDECODe method for OOD detection with a bounded false detection rate (FDR).  
    
    \item {\bf Experiments.} We demonstrate the efficacy of iDECODe on OOD detection over image and audio datasets, obtaining state-of-the-art (SOTA) results. We also show that iDECODe can be used for adversarial example detection.
\end{compactitem}

\textbf{\\ Related Work.}
There are a great deal of techniques for OOD detection, broadly in three categories:

\textbf{Supervised:} These techniques assume access to the OOD data or a proxy during the training phase of the detector.~\citet{kl-div} propose training the OOD detector with a predictive distribution following a label-dependent probability for the iD and a uniform distribution for the OOD datapoints. Similarly,~\citet{OE} propose training an OOD detector based on a distinct classification or density loss for the iD and OOD datapoints.~\citet{towardsNN} use a Bayesian framework for modeling iD and OOD datapoints separately.~\citet{mahalanobis} consider the Mahalanobis distance in the iD feature space to detect OOD datapoints. Logistic regression, trained on a small set of iD and OOD datapoints, is used to assign a score to the input by computing Mahalanobis distance of the noisy input from all layers of a classifier (trained for classification of the iD data). This score is expected to be higher for the iD than for OOD data.~\citet{residual-flow} extend this idea by replacing the Mahalanobis distance with a likelihood function learned from residual-flow models. \cite{guan2019prediction} develop methods for conformal classification and OOD detection based on learning classifiers to discriminate between the classes.~\citet{kaur2021all} propose an integrated approach based on ensemble of different scores (softmax, mahalanobis etc.) for OOD detection. 

\textbf{Self-supervised:} These techniques use a self-labeled dataset for OOD detection~\cite{geometric, GOAD, aux}. This dataset is created by applying transformations to the iD data and labeling the transformed data with the applied transformation. A classifier is trained for the auxiliary task of predicting the applied transformation on the self-labeled dataset. The error in the classifier's prediction of the applied transformation is used as a score to detect OOD-ness of an input. 

\textbf{Unsupervised:} These detection techniques use only the iD data for OOD detection.~\citet{baseline} propose using the maximum softmax score from a classifier trained on the iD data as the baseline method (SBP) for OOD detection. These scores are expected to be higher for the iD and lower for the OOD datapoints. ODIN~\cite{odin} was proposed as an enhancement to SBP by further separating these scores for iD and OOD datapoints after adding perturbations to the input and temperature scaling to the classifier's confidence. Recently,~\citet{macedo2021entropic} proposed replacing softmax scores with isomax scores and entropy maximization for OOD detection. Other unsupervised detection techniques based on the difference in the density estimates~\cite{mahmood2021multiscale}, energy scores~\cite{energy-based},
trust scores~\cite{trust-score}, likelihood ratio~\cite{likelihood-ratio}, activation path~\cite{gram} between the iD and OOD datapoints have been proposed for OOD detection.

Compared to supervised approaches, iDECODe does not require access to OOD data. It only requires the model to learn iD equivariance with respect to a set $G$ of transforms, which is a desirable property leading to the accuracy boost of classifiers~\cite{cnn_eq1, cnn_eq2}.
Data augmentation with $G$ during the training of a classifier on the iD data is one way to learn $G$-invariant classification of the iD data~\cite{data_aug, chen2020group}.
The auxiliary task of predicting the applied transformation on a self-labeled dataset also encourages the classifier to learn $G$-equivariant representations of the iD data~\cite{qi2019}. 

To our knowledge none of the above self-supervised and unsupervised methods provide any theoretical guarantees on OOD detection.~\citet{pac} provide PAC-style guarantees on the OOD detection aiming to minimize false detections. This approach is however supervised, as it requires OOD datapoints to train the detector which may not generalize to unseen OOD datapoints. Recently, there has been interest in unsupervised detection based on ICAD~\cite{vanderbilt, bates2021testing}.~\citet{vanderbilt} propose to use Martingale test~\cite{fedorova2012} on $p$-values from NCM based on either variational autoencoders (VAE) or deep support vector data description (SVDD) for OOD detection in time series data, where a batch of data is available for detection.
~\citet{bates2021testing} focus on problems that arise in conformal detection when multiple points are tested for OOD-ness. iDECODe proposed for the detection of a single point as OOD is also built on ICAD framework, which guarantees a bounded false detection rate (FDR).




\section{Background}
\label{sec:back}
\textbf{Equivariance.} For a set $X$, a function $f$ is equivariant with respect to a set of transformations $G$, if there is an explicit relationship between the transformation $g \in G$ of the function input and a corresponding transformation $g'$ 
of its output\footnote{Equivariant maps, also known as intertwiners, or homomorphisms between representations, are studied in representation theory, e.g., in Schur's lemma from 1905 \cite{hall2015lie}.}:
    \begin{equation}
        f(g(x)) = g'(f(x)), \forall x \in X.
    \label{equivariance}
    \end{equation}
Invariance is a special case of equivariance, where $g'$ is the identity function, so the output is unaffected by the transformation $g$ of the input. 

\textbf{\\ Inductive conformal prediction and inductive conformal anomaly detection.} Conformal prediction (CP)~\cite{icp,cp} aims to test if a new datapoint $x_{l+1}$ conforms to the set $X=\{x_i : 1 \leq i \leq l\}$ of the training datapoints, quantitatively measured by a non-conformity measure (NCM). Some examples of previously proposed NCMs are based on the $k$-nearest neighbours algorithm~\cite{icp, dknn}, ridge regression~\cite{icp}, support vector machines~\cite{icp}, random forests~\cite{ncm-rf}, VAE and SVDD~\cite{vanderbilt}.

An NCM is a real-valued function $\mathcal \mathcal{A}(X,x_{l+1})$, that assigns a non-conformity score $\alpha_{l+1}$ for $x_{l+1}$ relative to datapoints in the training set $X$. The  score $\alpha_{l+1}$ indicates how different $x_{l+1}$ is relative to $X$. Higher $\alpha_{l+1}$ indicates that the new $x_{l+1}$ is more different from the training dataset. 

Conformal anomaly detection (CAD)~\cite{cad} uses this non-conformity score for the new datapoint $x_{l+1}$ to compute its $p$-value $p_{l+1}$: 
\begin{equation*}
    p_{l+1} = \frac{|\{i=1,...,l\}: \alpha_i \geq \alpha_{l+1}|+1}{l+1}.
\end{equation*}
Here $\{\alpha_i: i=1,...,l\}$ is the set of non-conformity scores for the training set computed from the new set composed of the training set and $x_{l+1}$.
If $p_{l+1}$ is smaller than a given anomaly threshold $\epsilon \in (0, 1)$, then $x_{l+1}$ is classified as a conformal anomaly~\cite{cad}.
CAD can be viewed as performing statistical hypothesis testing;  the null hypothesis is a new datapoint $x_{l+1}$ and training datapoints $x_i$ for $1 \le i \le l$ are IID, and it is tested at the significance level $\epsilon$. Under the null, the probability of rejection is at most $\epsilon$~\cite{icad_hypothesis_testing}.

A drawback of the CAD framework is that it can be computationally inefficient. For every new datapoint, re-calculating $\alpha_i$ for all the training datapoints relative to the new set containing this datapoint might become computationally infeasible if evaluating the underlying NCM $\mathcal \mathcal{A}$ is expensive. To resolve this issue, inductive conformal anomaly detection (ICAD)~\cite{icad} was introduced based on the inductive conformal prediction (ICP) framework~\cite{icp}. Here, the training set is split into a \textit{proper training set} $X_{\text{tr}} = \{x_i: i=1,\ldots,m\}$ and a \textit{calibration set} $X_{\text{cal}}=\{x_j: j=m+1,\ldots,l\}$. Further, the non-conformity scores for each datapoint $x_j \in X_{\text{cal}}$ and the new datapoint $x_{l+1}$ are calculated relative to the proper training set:
\begin{equation*}
    \alpha_j = \mathcal \mathcal{A}(X_{\text{tr}}, x_j)
    \qquad \text{and} \qquad
    \alpha_{l+1} = \mathcal \mathcal{A}(X_{\text{tr}}, x_{l+1}).
\end{equation*}
ICAD uses these non-conformity scores to compute the $p$-value of $x_{l+1}$
\begin{equation*}
    p_{l+1} = \frac{|\{ j=m+1,...,l  : \  \alpha_{j} \geq \alpha_{l+1}\}|+1}{l-m+1}.
\label{icad}
\end{equation*}
Again, $x_{l+1}$ is classified as conformal anomaly if $p_{l+1}<\epsilon$.

\section{OOD Detection with Conformal Prediction}
\label{sec:tech}
We propose to use ICAD for OOD detection with a model trained to learn iD equivariance with respect to a set $G$ of transformations. Here, we first define a novel NCM and non-conformity score and then formalize iDECODe's algorithm for OOD detection with a bounded false detection rate.

\subsection{Novel base and aggregated NCMs}

\textbf{The proposed base NCM.} For an input $x$ and a transformation $g \in G$, we define a novel NCM (that we also refer to as a ``base NCM'')  as the error in the expected behavior of the transformation-equivariance learned by a model $M$ for the transformations $G$ on the proper training set $X_{\text{tr}}$:
\begin{align}
    \!\! \mathcal\mathcal{A}(X_{\text{tr}}, x; g) \! := \!\mathrm{Error}(M, x, g) \! = \! L[M(g(x)), g'M(x)]. \!
\label{eqn:ncscore}
\end{align}
Here, $L$ is a loss function, and recall that $g'$ is an ``output transform'' that depends on $g$. 

\textbf{Example NCM  based on data augmentation.} 
Data augmentation with $G$ during the training of a classifier on the iD data is used to learn invariant $(g'=I)$ classification of the iD data~\citep{baird1992document,data_aug,chen2020group,chatzipantazis2021learning}. Non-conformance in the label prediction between the original and the transformed input can thus be used as the base NCM:
\begin{equation}
\label{non_conf_loss}
    L[M(g(x)), g'M(x)] = \|M(g(x))-M(x)\|_2^2.
\end{equation}
The choice of the loss can be significant. For instance, the KL-divergence may be low both in- and out-of-distribution, for different reasons. For iD data, it can be low because the model learned the correct equivariance, while for OOD data it may be low because the predicted distribution is close to uniform for both the original and transformed OOD datapoints\footnote{Some existing OOD detection techniques~\citep{baseline, kl-div} assume that the softmax distribution for OOD datapoints is close to the uniform distribution.}. In our experiments, KL-divergence of the softmax scores as the base NCM lead to a relatively poor performance.

\textbf{Example NCM based on auxiliary task.} One can add the objective (or auxiliary task) of predicting a transformation $g \in G$ applied to the iD datapoint to ``encourage'' learning $G$-equivariant representations of the iD data~\citep{qi2019, geometric, aux, GOAD}. The error in the prediction of the transformation can thus be used as the base NCM:
\begin{equation*}
    L[M(g(x)), g'M(x)] = L[M(g(x)), g].
\end{equation*}
Here we formally set $g'$
such that its action $g'M(x)$ on any input $M(x)$ simply equals $g$, which is a special type of equivariance where the output does not depend on the input.   
Here, $L[M(g(x)), g]$ could be $\|M(g(x)) - g\|_2^2$ if $G$ is the set of parameterized transformations (such as affine or projective transformations) and $M(g(x))$ predicts the parameters of $g$. For discrete transformations, $L[M(g(x)), g]$ could be $\mathrm{Cross Entropy Loss}(M(g(x)), g)$.

\textbf{\\
The proposed aggregated non-conformity score.} Instead of using a single transformation $g$ as in the expression for $\mathcal \mathcal{A}$, we propose combining scores corresponding to several transformations. The intuition is that a single $\mathcal \mathcal{A}(X_{\text{tr}},x, g)$ might provide only noisy information of the OOD-ness of the input. By combining information over multiple transformations, we may reduce this noise, as it is less likely for OOD datapoints to behave as iD samples under multiple transformations.

Given $n$ transformations $g_{1:n} = (g_{1},\ldots, g_{n})$, we define the vector of base NCMs as
\begin{align}\label{v}
\begin{split}
    \mathcal{V} (x, X_{\text{tr}}; g_{1:n}) :=&  \left(\mathcal \mathcal{A}(X_{\text{tr}}, x; g_1), \ldots, \mathcal \mathcal{A}(X_{\text{tr}}, x; g_n)\right).
\end{split}
\end{align}
Our final non-conformity score is obtained by applying an aggregation function $F:\mathbb{R}^n \to \mathbb{R}$ to the vector $\mathcal{V}$ to aggregate the coordinate scores. We define $F(\mathcal{V} (x, X_{\text{tr}}; g_{1:n}))$ as the \emph{aggregated NCM}. It will always be clear from the context whether we refer to the base or aggregated NCM.

The only requirement for $F$ is that Theorem \ref{thm:justification_F} showing the correctness of the resulting $p$-value should apply. In particular, if the distribution of the data has a density, it is sufficient if $F$ is a coordinate-wise increasing  function in the non-conformity scores.  In our experiments we use 
$F(\mathcal{V}(x, X_{\text{tr}}; g_{1:n}))=\sum_{i=1}^n \mathcal \mathcal{A}(X_{\text{tr}}, x; g_i).$

Our base NCM $\mathcal \mathcal{A}$ and score vector $\mathcal{V}$ are defined for fixed $g$ and $g_{1:n}$ respectively. However, to ensure that the resulting scores are valid for use in conformal inference, we need the scores to be exchangeable random variables. For this, we will randomly and independently sample all transformations from some distribution $Q_G$ over $G$
\footnote{This requires a $\sigma$-algebra on $G$, which is used tacitly in our work with details given in Appendix D.} 
for computing the $p$-value.


\subsection{Algorithm and guarantee for OOD detection} \label{sec:theory}

A key theoretical result guarantees that the conformal inference framework applies to this construction. Compared with standard conformal prediction \citep{icp,cp}, we need to make sure that the 
exchangeability of scores still holds in the presence of randomness in the transforms.
For simplicity, in the next result, we assume that ties between $F(\mathcal V(x))$ and $F(\mathcal V(x_j))$ occur with zero probability. This holds under a broad set of practical assumptions, for instance if the distribution of the data is absolutely continuous with respect to Lebesgue measure (and thus has a density), and if $F$ is coordinate-wise strictly increasing. If this does not hold, then we can use smoothed $p$-values as in \cite{icp}. Although our experimental results are reported assuming no ties between $F(\mathcal V(x))$ and $F(\mathcal V(x_j))$, we obtained essentially the same results with smoothed $p$-values.

\begin{theorem} \label{thm:justification_F}
Let $G$ be a set of transformations. For each datapoint $x_j$ in the calibration set $X_\text{cal}$, let
\begin{equation*}
    \mathcal{V} (x_j) =  \mathcal{V} (x_j, X_\text{tr}; g_{j1},\ldots, g_{jn}) \text{ as defined in \eqref{v}},
\end{equation*}
where for each $i=1,\ldots,n$, $g_{ji}$ is sampled independently from some distribution $Q_G$ over $G$.  Given a test datapoint $x$, let  $\mathcal{V} (x) =  \mathcal{V} (x, X_{\text{tr}}; g_{x1},\ldots, g_{xn})$ as in \eqref{v}, where for $i=1,\ldots,n$, $g_{xi}$ is also sampled independently from $Q_G$. If $x$ is in the training distribution $D$, then for any $F:\mathbb{R}^n \to \mathbb{R}$, the $p$-value of $x$
\begin{equation}
  P = \frac{|\{j=m+1,\ldots,l:F(\mathcal V(x_j)) \ge F(\mathcal V(x))\}| + 1}{l-m+1} 
\label{p-value}
\end{equation}
is uniformly distributed over $\{1/(k+1), 2/(k+1),...,1\}$, where $k=l-m$.
\end{theorem} 
See Appendix A
for a proof.
Under the null hypothesis that an input datapoint $x$ is from the training distribution (i.e., $x\sim D$), and assuming ties occur with zero probability, the $p$-value  from equation~\ref{p-value} is distributed uniformly over $\{1/(k+1),\ldots, 1\}$, where $k=m-l$. We can detect if a datapoint is OOD by rejecting the null; if the $p$-value falls below a detection threshold $\epsilon \in (0, 1)$, i.e., $P < \epsilon$, then $x$ is an OOD data  point (see Algorithm~\ref{alg:ood_det}).\footnote{We remark that in statistics, it is common to use $P\le \ep$ for rejecting a null hypothesis \citep{lehmann1998theory}; however here we follow \citet{icad} to write $P<\ep$.}
The next result states that the false detection probability is bounded; this is a consequence of standard results on conformal inference~\citep{cp}, and we provide the proof only for completeness in Appendix B.
\begin{algorithm}[t!]
   \caption{iD equivariance for conformal OOD detection}
   \label{alg:ood_det}
\begin{algorithmic}
   \STATE {\bfseries Input:} test datapoint $x$, model $M$ trained on proper training set, distribution $Q_G$ over transforms, loss function $L$ for equivariance, vectors of scores $\{\mathcal V(x_j): j=m+1,\ldots,l\}$ over calibration set,  aggregation function $F$, desired false positive rate $\epsilon \in (0, 1)$
   \STATE {\bfseries Output:} "$1$" if $x$ is detected as OOD; "$0$" otherwise
   \STATE $(g_1, \cdots, g_n) \sim Q_G^n$
   \STATE $\mathcal V(x) = \{L[M(g_i(x)), g'_iM(x)] \mid 1 \le i \le n\}$ 
   \STATE Set $p$-value $P = \frac{|\{j=m+1,\ldots,l:F(\mathcal V(x_j)) \ge F(\mathcal V(x))\}| + 1}{l-m+1}$
   \STATE \textbf{Return} $\mathbbm{1}(P < \epsilon)$
\end{algorithmic}
\end{algorithm}

\begin{corollary}\label{cor:bound}
The probability of false OOD detection by iDECODe is upper bounded by $\epsilon$.
\end{corollary}
Thus, if the test datapoint $x$ and the datapoints in the training set are IID, then $\epsilon$ is an upper bound on the probability of detecting $x$ as OOD~\citep{icad}.

\section{Experimental Results}

Score from the proposed base NCM can be used to indicate the OOD-ness of an input. Higher score indicates more OOD-ness of an input. We use these scores for OOD detection and refer to it as the \textit{base score} method in our experiments. We will use the shorthand ICAD for running ICAD with the proposed base NCM for OOD detection, i.e., ICAD is iDECODe with $|\mathcal V(x)|=1$. For all experiments, $g_i$ for $i=1,\ldots,n$ is sampled independently from the uniform distribution $Q_G$ over $G$. All the reported results for base score method, ICAD and iDECODe are averaged over five runs with random sampling of the set of transformations.

\label{sec:exp}
\subsection{Results on vision datasets}
We first describe the model $M$ used in the experiments. Then, we illustrate the efficacy of iDECODe with results on ablation studies and comparison with SOTA results.

\textbf{\\ Learning $G$-equivariant iD representations from the proper training set.} Recently,~\citet{qi2019} proposed ``Autoencoding Variational Transformations'' (AVT) for learning transformation equivariant representations via  VAE.
Given a dataset $X$ and the set $G$ of transformations, AVT trains a VAE to predict the transformation $g \in G$ applied to an input $x\in X$.
They argue that the AVT model learns an encoded representation of $X$ that is $G$-equivariant by maximizing mutual information between the encoded space and $G$. We use an AVT model $M$ trained to be $G$-equivariant on the proper training set of the iD data for results on vision datasets.

\subsubsection{Ablation studies.}
Following the experimental conventions for OOD detection on vision datasets~\cite[see e.g.,][]{mahalanobis, aux}, we perform ablation studies on CIFAR-10 as the iD dataset and SVHN, LSUN, ImageNet, CIFAR100, and Places365 datasets as OOD. The details of these datasets and the AVT model $M$ with ResNet architecture trained to be $G$-equivariant on the proper training set (90\% of the total training data) of CIFAR-10 (with the same set of hyperparameters from~\citet{qi2019}'s AVT model on CIFAR-10) are given in Appendix C.1.1.

\textbf{The set $G$ and base NCM.} Since we use ~\citet{qi2019}'s AVT model trained on CIFAR-10 to perform our ablation studies, we use the same set $G$ of projective transformations used by their AVT model to learn the equivariance on CIFAR-10.  A projective transformation $g \in G$ is composed of scaling by a factor of $[0.8, 1.2]$, followed by a random rotation with an angle in $\{0^{\circ}, 90^{\circ}, 180^{\circ}, 270^{\circ} \}$, and random translations of the four corners in the horizontal and the vertical directions by a fraction of up to $\pm 0.125$ of its height and width. A projective transformation is represented by a matrix in $\mathbb{R}^{3\times 3}$. We therefore use mean square error between the parameters of the applied and predicted transformations as the base NCM, i.e. $L[M(g(x)), g'M(x)] = \|M(g(x)) - g\|_2^2$. Here $M(g(x))$ and $g$ are the vectors with parameter values of the predicted and applied transformation respectively.

\begin{figure*}[t!]

    \centering
    
    \begin{subfigure}
        \centering
        \includegraphics[height=0.85in]{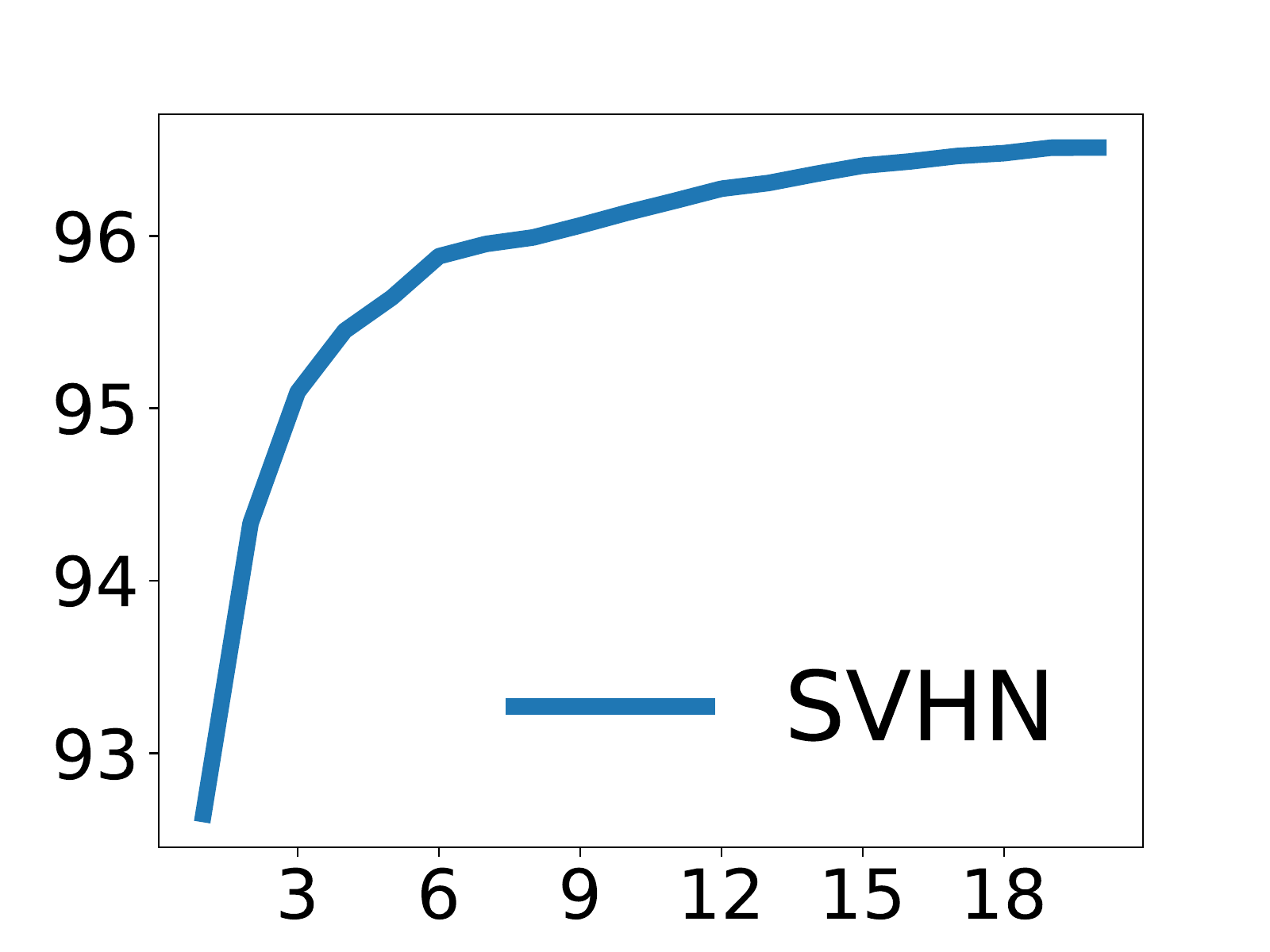}
    \end{subfigure}%
    \begin{subfigure}
        \centering
        \includegraphics[height=0.85in]{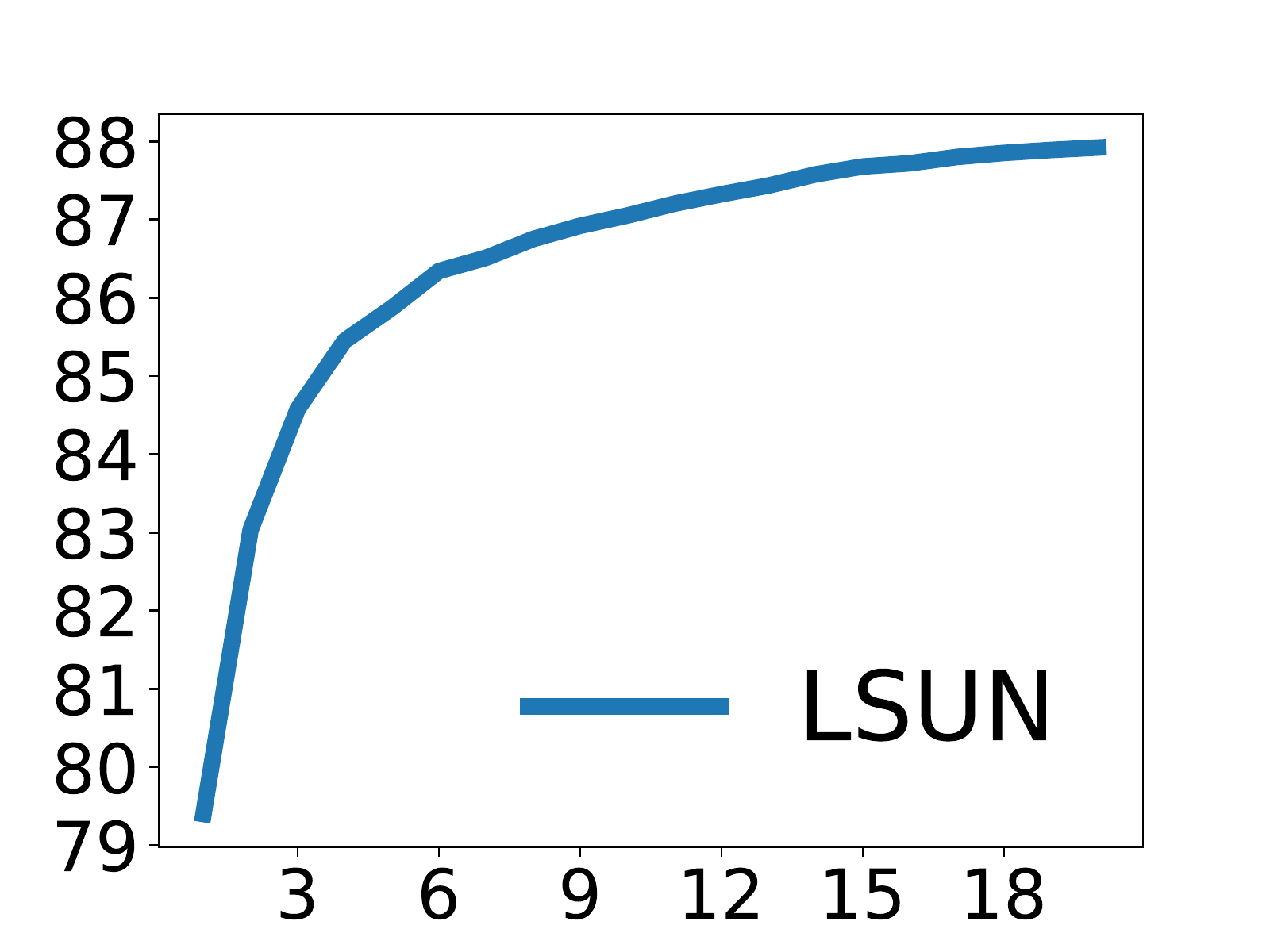}
    \end{subfigure}
    \begin{subfigure}
        \centering
        \includegraphics[height=0.85in]{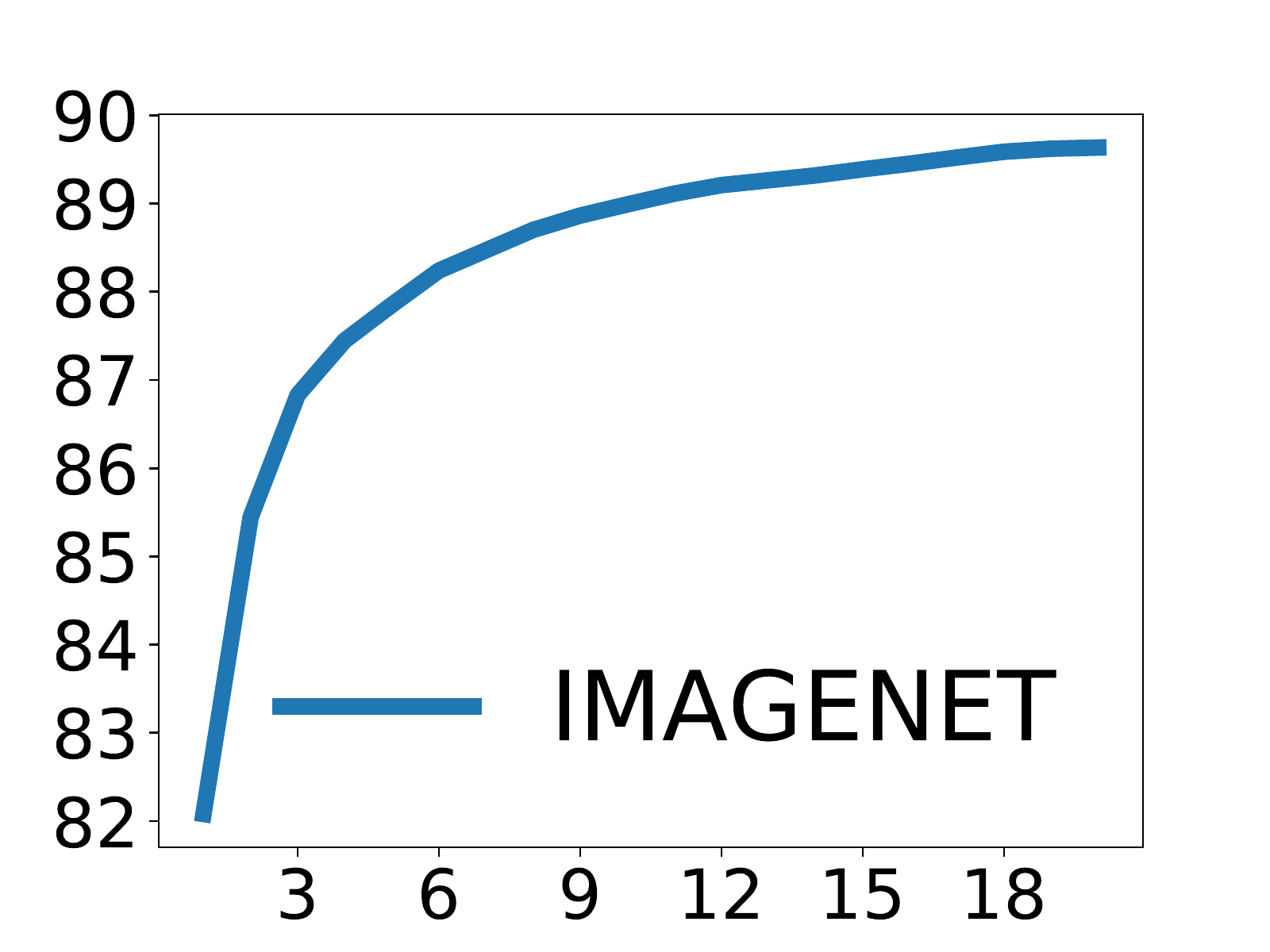}
    \end{subfigure}
    \begin{subfigure}
        \centering
        \includegraphics[height=0.85in]{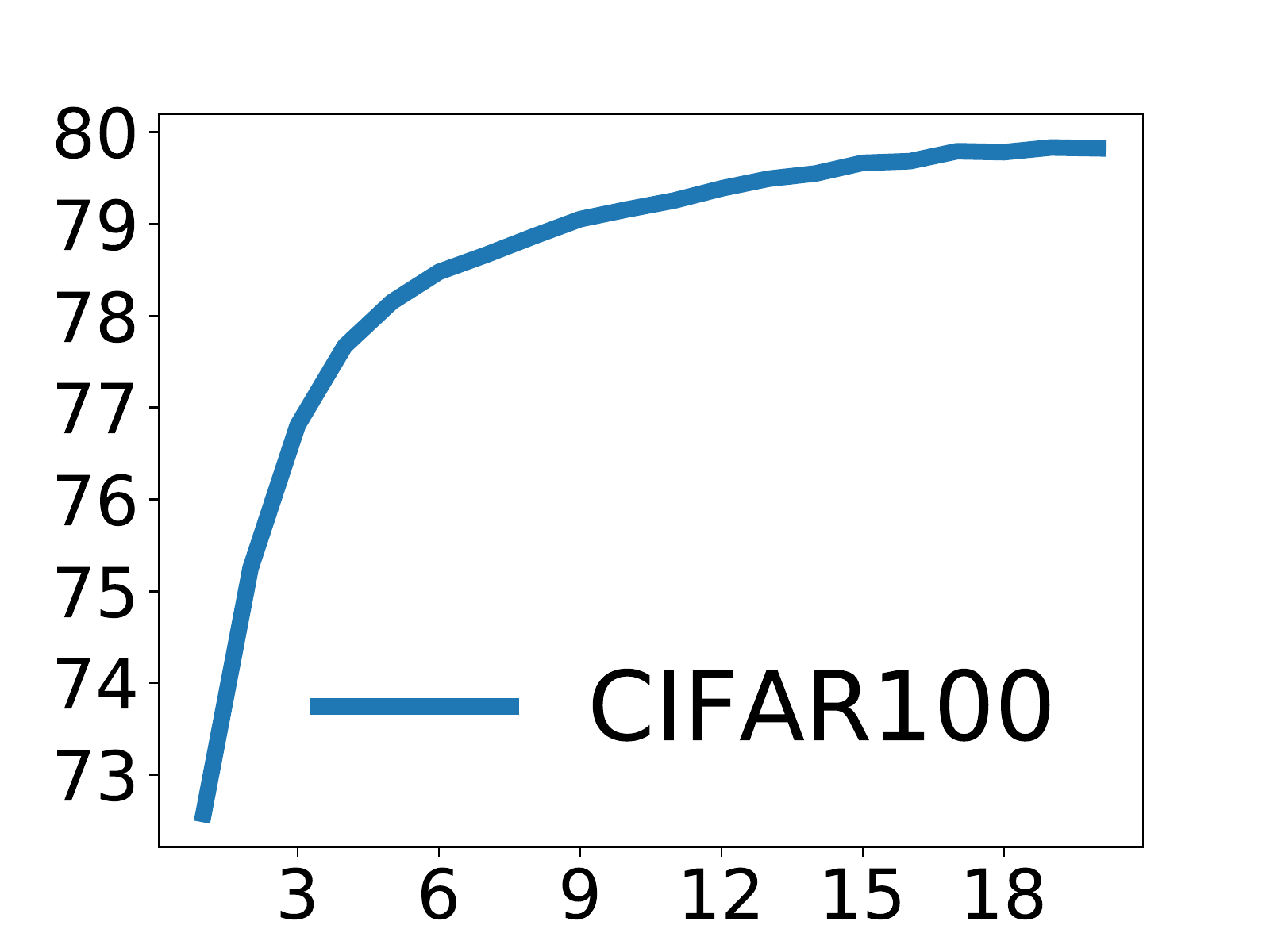}
    \end{subfigure}
    \begin{subfigure}
        \centering
        \includegraphics[height=0.85in]{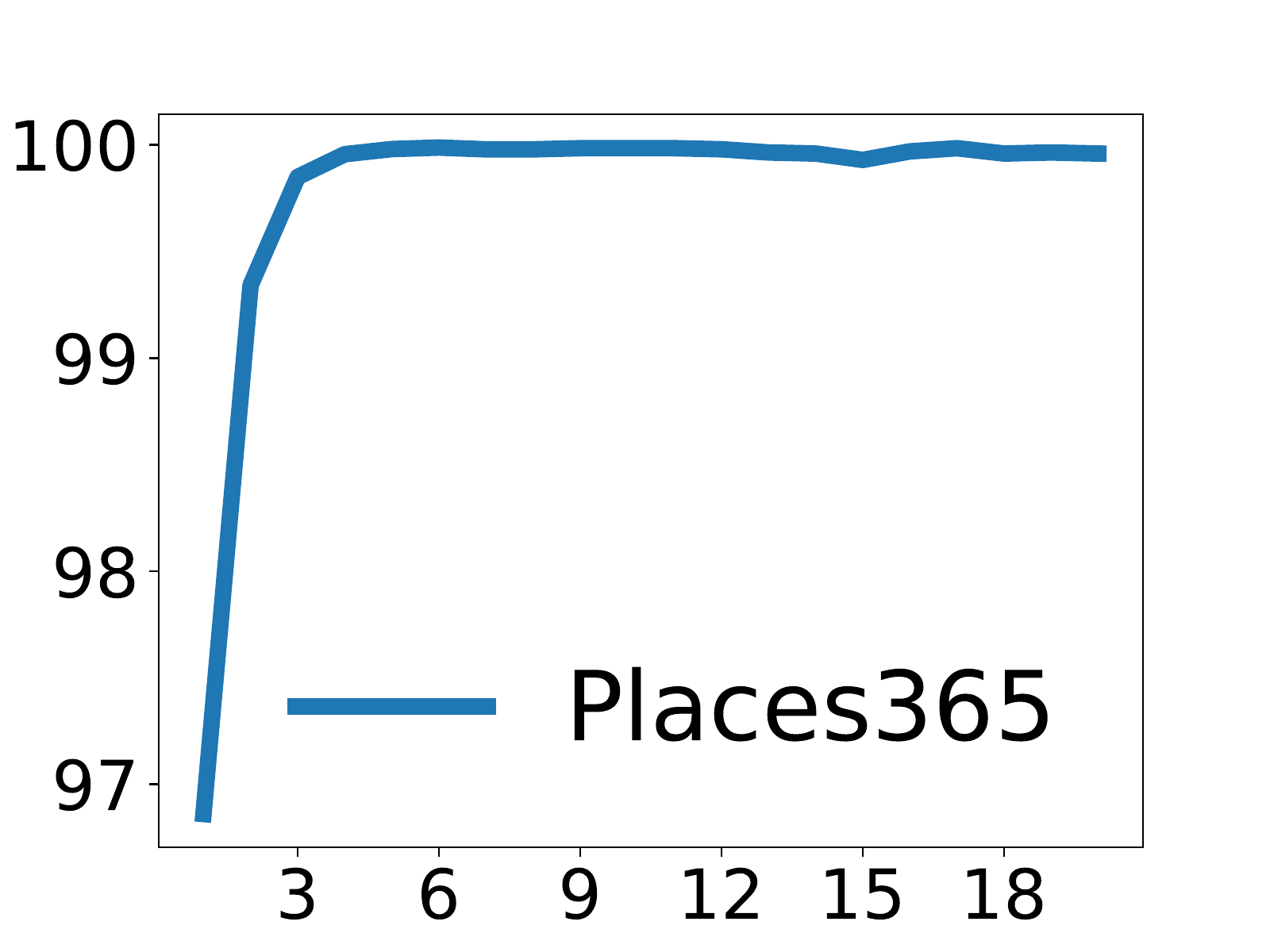}
    \end{subfigure}
 
    \caption{\footnotesize{AUROC vs $|\mathcal V(x)|$ reported by iDECODe on CIFAR-10 as iD.}}
    \label{fig:tnr_auroc_n}
\end{figure*}

\textbf{Results.} We call iD datapoints positives and OOD datapoints negatives. We use the True Negative Rate (TNR) at 90\% True Positive Rate (TPR), and the Area under Receiver Operating Characteristic curve (AUROC) for evaluation. The details of these metrics are given in Appendix C.
We illustrate the effectiveness of iDECODe by performing the following three ablation studies:

\begin{enumerate}[a)]\leftmargin0cm
\item \label{(a)} \textit{Comparison with (1) base score method and (2) ICAD.} 
Table~\ref{tab:comp_base_icad} shows that iDECODe with $|\mathcal V(x)|=5$ outperforms both the base score method and ICAD on all OOD datasets for both TNR and AUROC. 
Although the performance of base score method and ICAD is quite similar, ICAD detects OOD datapoints with a bounded FDR.
    
\item \label{(b)} \textit{Performance of iDECODe versus $|\mathcal V(x)|$.} Figure~\ref{fig:tnr_auroc_n} shows that AUROC increases with the size of $\mathcal V(x)$ ($|\mathcal V(x)| = 1,\ldots, 20$). Similar performance gain of iDECODe on TNR (90\% TPR) with  higher $|\mathcal V(x)|$ is reported in Appendix C.1.2. This justifies the intuition that it is unlikely for an OOD to behave as iD under multiple transforms in $G$ for which $M$ is iD-equivariant.
    
\item \textit{Controlling the False Detection Rate.} Figure~\ref{box_plot} shows that the FDR of iDECODe with $|\mathcal V(x)|=5$ is upper bounded by $\epsilon$ on average. These plots are generated with $\epsilon=0.05\cdot k$, $k=1,\ldots,10$. Calibration set of size 1000 images is randomly sampled with replacement from a held-out (not used in training) set of 5000 images. This is repeated five times, and we show box plots, with the median and inter-quartile range. Similar results on other sampled calibration set sizes are in 
Appendix C.1.2.
\end{enumerate}
We also perform ablation studies on SVHN as the iD dataset and all other datasets as OOD. The results are similar to the results on CIFAR-10 as iD, and reported in Appendix C.1.3.

\begin{table*}[t!]
	\begin{minipage}{0.5\linewidth}
\caption {\label{tab:comp_base_icad}\footnotesize{Comparison of base score, ICAD (iDECODe with $|\mathcal V(x)|=1)$ with ours (iDECODe with $|\mathcal V(x)|=5)$ on CIFAR-10 as iD.}}
\setlength{\tabcolsep}{3pt}
\begin{adjustbox}{width=1.2\columnwidth}
\begin{tabular}{c|ccc|ccc}
\hline
$D_{out}$  & \multicolumn{3}{c|}{TNR (90\% TPR)}  &   \multicolumn{3}{c}{AUROC}     \\ 
\cline{2-4} \cline{5-7}  
        & Base Score     & ICAD  & Ours                                                         
	&  Base Score     & ICAD   & Ours                                              \\ 
\hline
SVHN  &  84.18 $\pm$ 0.88  & 84.11 $\pm$ 0.89  & \textbf{93.81 $\pm$ 0.38} 
    & 92.86 $\pm$ 0.16 & 92.86 $\pm$ 0.16 & \textbf{95.70 $\pm$ 0.07}\\ 

LSUN &  55.70 $\pm$ 0.94 &   55.66 $\pm$ 0.92  & \textbf{64.22 $\pm$ 0.68}
    & 79.47 $\pm$ 0.26  & 79.47 $\pm$ 0.26 & \textbf{85.98 $\pm$ 0.13}\\
   
ImageNet &  60.76 $\pm$ 0.72  &  60.73 $\pm$ 0.70  & \textbf{70.29 $\pm$ 0.61}
    & 82.21 $\pm$ 0.18  & 82.21 $\pm$ 0.18& \textbf{87.97 $\pm$ 0.15 }\\

CIFAR100 & 43.82 $\pm$ 0.94  & 43.764 $\pm$ 0.89   &   \textbf{48.70 $\pm$ 0.39}
    & 72.61 $\pm$ 0.30 & 72.60 $\pm$ 0.30 & \textbf{78.04 $\pm$ 0.20}\\

Places365 &  88.58 $\pm$ 0.34  & 88.58$\pm$ 0.34 &  \textbf{99.97 $\pm$ 0.01}
   & 96.87 $\pm$ 0.05  & 96.87 $\pm$ 0.06 & \textbf{99.98 $\pm$ 0.01}\\
\hline
\end{tabular}

\end{adjustbox}
	\end{minipage}\hfill
	\begin{minipage}{0.35\linewidth}
		\centering
		\includegraphics[height=1.35in]{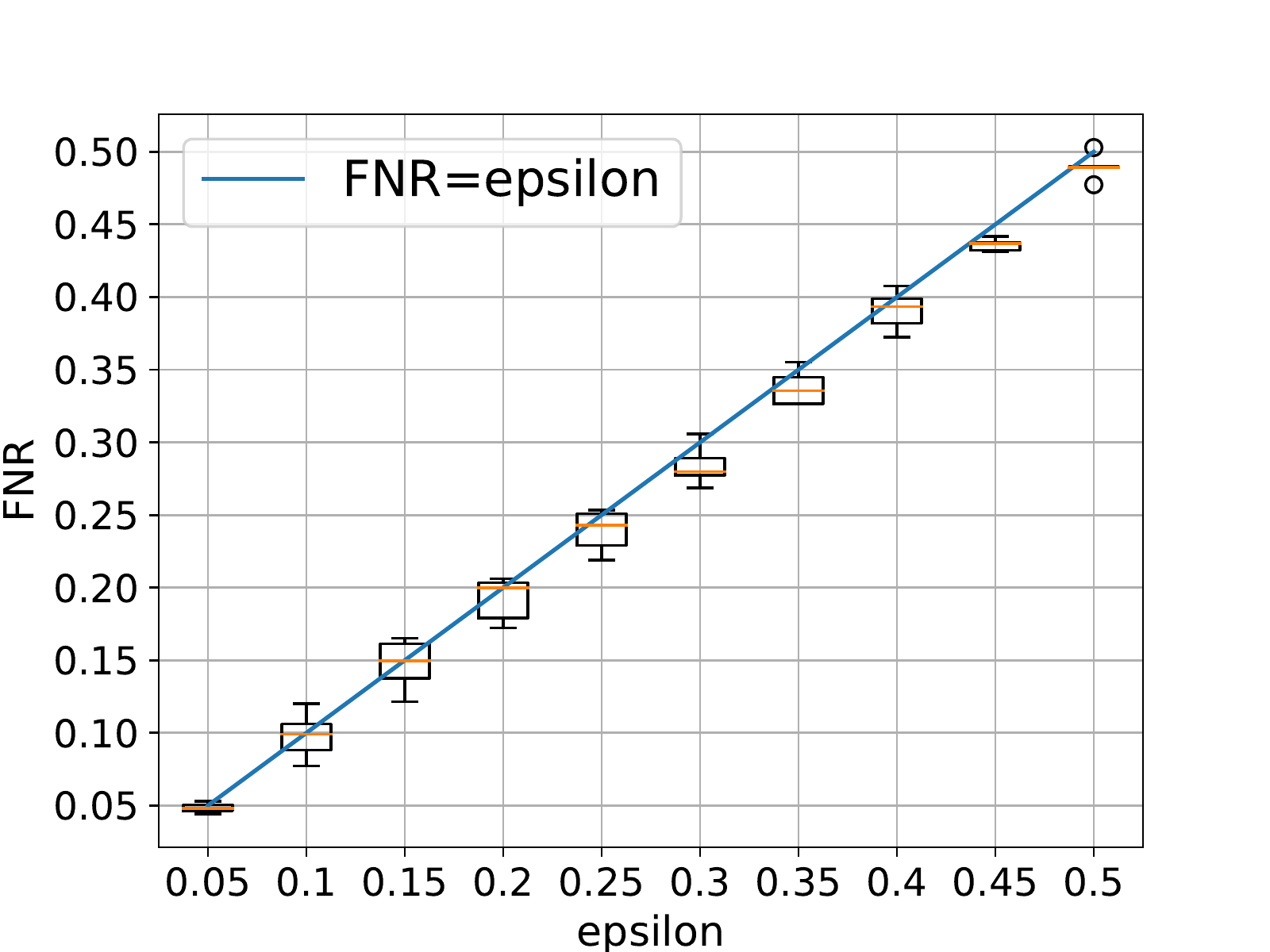}
		\captionof{figure}{\footnotesize{False detection rate of iDECODe is upper bounded by $\epsilon$ on average for $|X_{\text{cal}}|$ = 1000.
		\label{box_plot}
    }}
	\end{minipage}
\end{table*}

\subsubsection{Comparison with state-of-the-art.}
Here, we compare the performance of iDECODe with current SOTA self-supervised and unsupervised OOD detectors. Following the convention of one-class OOD detection on CIFAR-10 with WideResNet model (WRN)~\cite{aux,GOAD}, we also train the AVT model with WRN architecture on the proper training set (90\% of the total training set) of one class of CIFAR-10 (e.g., class ``0"). All other classes (e.g., classes ``1" to ``9") are considered as OOD. We also evaluate iDECODe on the one-class OOD detection for the 20 super-classes of CIFAR-100. Existing results for one-class OOD detection on CIFAR-100 have been reported on ResNet-18 model~\cite{csi}. Therefore, we also train the AVT model with ResNet-18 architecture on the proper training set (90\% of the total training set) of CIFAR-100. Model details for both CIFAR-10 and CIFAR-100 are given in Appendix C.1.4.


\textbf{The set $G$ and the base NCM.}
For a fair comparison with SOTA, we chose
$G$ to be the set of four classes of rotations: rotation by an integer-valued angle in the $[-10^{\circ}, 10^{\circ}]$, $[80^{\circ}, 100^{\circ}]$, $[170^{\circ}, 190^{\circ}]$ and $[260^{\circ}, 280^{\circ}]$ range. This is because the current SOTA~\cite{aux} uses prediction of the applied rotation angle from the set of four rotation angles ($\{0^{\circ}, 90^{\circ}, 180^{\circ}, 270^{\circ}\}$) as the auxiliary task for OOD detection. Since the prediction of the AVT model (trained to learn equivariance with respect to rotation angle ranges) is the softmax scores for the four classes of rotation angle ranges on the input, we use $L[M(g(x)), g'M(x)]= \mathrm{Cross Entropy Loss}(M(g(x)), g)$ as the base NCM. Here $g$ is the one-hot vector with ``one" for the applied class and ``zero" for the other three, and $M(g(x))$ is the vector of softmax scores of the four classes predicted by $M$. 


\textbf{Results and methods.} Table~\ref{sota-results} shows the comparison of AUROC between the existing SOTA methods, ICAD and iDECODe with $|\mathcal V(x)|=5$ on the one-class OOD detection of CIFAR-10 (left) and CIFAR-100 (right). For CIFAR-100, we report the mean AUROC across all classes. Results for individual super classes of CIFAR-100 have been included in Appendix C.1.4. We achieved SOTA on five classes and the overall mean for CIFAR-10. On CIFAR-100, we achieved competitive results on most of the classes and the mean.

One-class SVM (SVM)~\cite{oc-svm} and DeepSVDD (SVDD)~\cite{deep-svdd} are unsupervised methods, modeling the training distribution as a single class. Points outside of the iD class are detected as OOD datapoints. All other techniques are self-supervised, training a classifier with an auxiliary task of predicting the applied transformation. The error in the prediction is used for OOD detection. GM~\cite{geometric} uses a set of geometric transformations. GOAD~\cite{GOAD} generalizes GM to a class of affine transformations. RNet~\cite{rotnet} uses the set of $\{0^{\circ}, 90^{\circ}, 180^{\circ}, 270^{\circ}\}$ rotations.
AUX~\cite{aux} predicts the applied rotation, and the vertical and horizontal translations separately. The error in the three predictions is summed for OOD detection. CSI~\cite{csi} uses distributionally-shifted augmentations of the iD data as negative samples in the contrastive learning framework to differentiate iD and OOD datapoints. An auxiliary task of predicting the applied shifting transform improves the features learnt by contrastive learning. Score functions on these features are used for detection.

The results of the existing techniques on CIFAR-10 are as reported in~\citet{aux}, while for GOAD they are taken from their paper~\cite{GOAD}. For CSI, results reported on CIFAR-10 in their paper are on ResNet-18 architecture, so we generate and report CSI results for one-class OOD detection of CIFAR-10 on the same WRN architecture as used for other methods. For CIFAR-100, the results of the existing techniques are as reported in~\citet{csi}. Although we could not out-perform CSI on most of the CIFAR-100 classes, iDECODe does not require access to any OOD data during training. CSI on the other hand uses distributionally shifted augmentations of iD data as OOD data for contrastive learning of iD (or positive) and OOD (or negative) datapoints. Also, to our knowledge, CSI does not provide any detection guarantees.



\begin{table*}[!t]
\caption {\label{sota-results}\footnotesize{AUROC for one-class OOD detection on CIFAR-10 (left) and mean AUROC across all classes for one-class OOD detection on CIFAR-100 (right) by SOTA (SVM,$\ldots$,CSI), ICAD($|\mathcal V(x)|=1$) and Ours($|\mathcal V(x)|=5$).
Best results are in bold and second best are underlined.}}
\setlength{\tabcolsep}{3pt}
\centering
\begin{adjustbox}{width=1.2\columnwidth}
\begin{tabular}{cccccccccc}
\toprule
Class &  SVM & SVDD & GM & RNet & GOAD & AUX & CSI & ICAD & Ours \\
\midrule
0    & 65.6 & 61.7 & 76.2 & 71.9 & 77.2 & 77.5 & 76.51 $\pm$ 0.12 & \underline{85.94 $\pm$ 0.09} & \textbf{86.46 $\pm$ 0.03} \\
1    & 40.9 & 65.9 & 84.8 & 94.5 & 96.7 & 96.9 &  \textbf{98.68 $\pm$ 0.02}  & 97.82 $\pm$ 0.09 & \underline{98.11 $\pm$ 0.02} \\
2    & 65.3 & 50.8 & 77.1 & 78.4 & 83.3 &  \underline{87.3} &  \textbf{88.30 $\pm$ 0.06}  & 86.14 $\pm$ 0.34 & 86.04 $\pm$ 0.46 \\
3    & 50.1 & 59.1 & 73.2 & 70.0 & 77.7 & 80.9 &  79.38 $\pm$ 0.08  & \underline{81.59 $\pm$ 0.23} & \textbf{82.57 $\pm$ 0.11} \\
4    & 75.2 & 60.9 & 82.8 & 77.2 & 87.8 & \textbf{92.7} & 88.15 $\pm$ 0.11 & 90.27 $\pm$ 0.13  & \underline{90.87 $\pm$ 0.05} \\
5    & 51.2 & 65.7 & 84.8 & 86.6 &  87.8 & \underline{90.2} &  \textbf{91.79 $\pm$ 0.07} & 88.57 $\pm$ 0.21 & 89.24 $\pm$ 0.12 \\
6    & 71.8 & 67.7 & 82.0 & 81.6 &  90.0 & \textbf{90.9} &  \underline{88.49 $\pm$ 0.04}&  88.10 $\pm$ 0.15 &  88.18 $\pm$ 0.40 \\
7    & 51.2 & 67.3 & 88.7 & 93.7 & 96.1 & 96.5 &  97.08 $\pm$ 0.02  & \underline{97.55 $\pm$ 0.09} & \textbf{97.79 $\pm$ 0.06} \\
8    & 67.9 & 75.9 & 89.5 & 90.7 & 93.8 & 95.2 &  96.00 $\pm$ 0.04  & \underline{96.96 $\pm$ 0.06} & \textbf{97.21 $\pm$ 0.03} \\
9    & 48.5 & 73.1 & 83.4 & 88.8 & 92.0 & 93.3 &  93.31 $\pm$ 0.09  &  \underline{95.29 $\pm$ 0.08} &\textbf{95.46 $\pm$ 0.10} \\
\midrule
Mean & 58.8 & 64.8 & 82.3 & 83.3 & 88.2 & 90.1 & 89.97 & \underline{90.82}  & \textbf{91.19} \\
\bottomrule
\end{tabular}
\end{adjustbox}
\quad
\begin{adjustbox}{width=0.3\columnwidth}
\begin{tabular}{ccccccccc}
\toprule
Method & AUROC \\
\midrule
SVM & 63.1 \\
GM    & 78.7 \\
AUX & 77.7 \\
AUX+Trans    & 79.8 \\
GOAD    & 74.5 \\ 
CSI     & \textbf{89.6} \\
ICAD    &   80.31 \\
Ours & \underline{80.66} \\
\bottomrule
\end{tabular}
\end{adjustbox}
\end{table*}

\begin{table*}
\caption {\label{tab:comp_sbp}Comparison with~\citet{baseline}'s SOTA method (SBP) on CIFAR-10. Best results are in bold.} 
\setlength{\tabcolsep}{3pt}
\centering
\begin{adjustbox}{width=1.6\columnwidth}
\begin{tabular}{c|ccc|ccc}
\hline
$D_{out}$  & \multicolumn{3}{c|}{TNR (90\% TPR)}  &   \multicolumn{3}{c}{AUROC}     \\ 
\cline{2-4} \cline{4-7}  
        & SBP \textbf{(SOTA)}      & Ours ($|(\mathcal V(x)=5)|$)    & Ours ($|(\mathcal V(x)=20)|$)                                                    
	&  SBP       & Ours ($|(\mathcal V(x)=5)|$)    & Ours ($|(\mathcal V(x)=20)|$)                                        \\ 
\hline
SVHN  &  55.71 & 93.81 $\pm$ 0.38 & \textbf{97.12 $\pm$ 0.13} 
    &87.86 & 95.70 $\pm$ 0.07 & \textbf{96.50 $\pm$ 0.01}\\ 
LSUN &    66.14  & 64.22 $\pm$ 0.68 & \textbf{67.54 $\pm$ 0.34}
    &   \textbf{89.90} & 85.98 $\pm$ 0.13 & 87.94 $\pm$ 0.03\\
ImageNet &   59.13 & 70.29 $\pm$ 0.61 & \textbf{73.67 $\pm$ 0.45}
    & 87.32 & 87.97 $\pm$ 0.15  & \textbf{89.58 $\pm$ 0.06}\\
CIFAR100 & 50.52   & 48.70 $\pm$ 0.39  &   \textbf{51.15 $\pm$ 0.23}
    & \textbf{83.13} & 78.04 $\pm$ 0.20 & 79.92 $\pm$ 0.09\\
Places365 & 52.77    & 99.97 $\pm$ 0.01 &   \textbf{100.0 $\pm$ 0.0}
   & 83.48  & 99.98 $\pm$ 0.01 & \textbf{99.97 $\pm$ 0.02}\\
\hline  
\end{tabular}
\end{adjustbox}
\end{table*}

\textbf{Comparison with SBP as SOTA. } We also compare with~\citet{baseline}'s OOD detection technique (SBP) based on maximum softmax score. Since SBP depends on the softmax score of the predicted class, it cannot be applied to the one-class OOD detection problem. We therefore compare the performance of SBP with CIFAR-10 as iD and all other datasets as OOD. Table~\ref{tab:comp_sbp} shows that
iDECODe (both with $\mathcal |V(x)| = 5 \text{ and} \mathcal |V(x)| = 20$) could outperform SBP in most cases. The training details of the classifier used in SBP method are provided in Appendix C.1.5.



For evaluating the generalizability of iDECODe across different transformation sets, we compare the performance of iDECODe on CIFAR-10 dataset as iD and others as OOD with two transformation sets (one with the projective transformations and the other with rotations). iDECODe with both transformation sets could perform reasonably well and these results are reported in Appendix C.1.6.

\subsection{Results on audio data}
\textbf{Dataset and iD equivariance.} FSDNoisy18k (FSD)~\cite{audio_dataset} is an audio dataset with verified (correctly labeled) and noisy data on twenty classes. We use only the verified subset of FSD. Based on similarity between classes (e.g., fire and fireworks) we group them into four sets, each containing five distinct classes. Details about these sets are included in Appendix C.2.
Similar to the vision experiments, we perform one-class OOD detection on the four sets (e.g., set 0 as the iD dataset and other three sets as OOD). We train a VGG classifier on the proper training data of the iD dataset to classify the classes in the set, using data augmentation to learn iD equivariance. Fifteen datapoints from each class in the iD training set are held out as calibration data for the set (this is the same setting as ~\citet{audio_ood}'s validation set in FSD).

\textbf{\\ The set $G$ and the base NCM.} We use the set $G$ of time and frequency masks proposed in SpecAugment~\cite{specaug} for data augmentation in speech recognition. We use non-conformance between the label predictions of the original and transformed inputs from~\eqref{non_conf_loss} as the base NCM. 

\textbf{\\ Results.}
Table~\ref{tab:audio_comp_softmax_non_conf} shows the results for AUROC on audio OOD detection. iDECODe with $|\mathcal V(x)|=20$ achieves SOTA results on all the sets. 

\begin{table*}[!t]
\caption {\label{tab:audio_comp_softmax_non_conf}\footnotesize{
AUROC for audio OOD detection by SBP as SOTA, base score method, ICAD and iDECODe.}} 
\setlength{\tabcolsep}{3pt}
\centering
\begin{adjustbox}{width=1.2\columnwidth,center}
\begin{tabular}{c|ccccc}
\hline
$D_{in}$       
        & SBP (\textbf{SOTA}) & Base Score  & ICAD($|\mathcal V(x)|=1$) & Ours ($|\mathcal V(x)|=5$) & Ours ($|\mathcal V(x)|=20$)                                                                                               \\ 
\hline
Set 0 & 59.39  & 61.46 $\pm$ 0.72 & 62.19 $\pm$ 0.26 & 65.07 $\pm$ 0.39 & \textbf{66.21 $\pm$ 0.39}\\ 
Set 1 &  53.77 & 52.88 $\pm$ 0.86 & 52.48 $\pm$ 1.72 & 54.69 $\pm$ 0.75 & \textbf{56.18 $\pm$ 0.27}\\
Set 2 & 58.04 & 57.97 $\pm$ 0.61 & 57.41 $\pm$ 0.81 & 60.91 $\pm$ 0.71 & \textbf{61.13 $\pm$ 0.18}\\
Set 3 & 50.07  & 55.18 $\pm$ 0.67 & 54.35 $\pm$ 1.38 & 56.56 $\pm$ 1.10 & \textbf{56.83 $\pm$ 0.34}\\
\midrule
Mean & 55.32 & 56.87 & 56.61 & 59.31 & \textbf{60.09}\\
\bottomrule
\end{tabular}
\end{adjustbox}
\end{table*}

\begin{table*}[!t]
\caption{\footnotesize{AUROC for adversarial detection on CIFAR-10 by supervised SOTA (KD+PU, LID, Mahala), unsupervised SOTA (Odds, AE), ICAD (iDECODe with $|\mathcal V(x)|=1$). and Ours (iDECODe with $|\mathcal V(x)|=5$). Best results are in bold and second best are underlined.}}
\label{table:adv_table1}
\setlength{\tabcolsep}{3pt}
\centering
\begin{adjustbox}{width=1.4\columnwidth}
\begin{tabular}{c|cccc|cccc}
\hline
Method & \multicolumn{4}{c|}{ResNet}  &   \multicolumn{4}{c}{DenseNet}     \\ 
\cline{2-5} \cline{5-9} 
        & FGSM     & BIM  & DF & CW & FGSM  & BIM  & DF & CW  \\

\hline

   KD+PU &   83.51 & 16.16 & 76.80 & 56.30 & 85.96 & 3.10 & 68.34 & 53.21  \\ 

   LID & \underline{99.69} & 95.38 & 71.86 & 77.53 & \underline{98.20} & 94.55 & 70.86 & 71.50\\

   Mahala &  \textbf{99.94} & \textbf{98.91} & 78.06 & \underline{93.90} & \textbf{99.94} & \textbf{99.51} & 83.42 & 87.95\\

   Odds & 46.32 & 59.85 & 75.58 & 57.58 & 45.23 & 69.01 & 58.30 & 61.29\\

   AE & 97.24 & 94.93 & 78.19 & 74.29 & 78.38 & \underline{97.51} & 65.31 & 68.15\\
   
   ICAD & 93.1 $\pm$ 0.1 & 92.6 $\pm$ 0.2 & \underline{92.6 $\pm$ 0.3} & 91.7 $\pm$ 0.1 & 92.3 $\pm$ 0.3 & 92.2 $\pm$ 0.2 & \underline{90.7 $\pm$ 0.2} & \underline{90.3 $\pm$ 0.2} \\
   Ours & 96.6 $\pm$ 0.1 & \underline{96.4 $\pm$ 0.1} & \textbf{96.4 $\pm$ 0.1} & \textbf{95.9 $\pm$ 0.1} & 96.2 $\pm$ 0.1 & 96.1 $\pm$ 0.1 & \textbf{95.2 $\pm$ 0.1} & \textbf{95.0 $\pm$ 0.1}\\
\hline
\end{tabular}
\end{adjustbox}
\end{table*}

\subsection{Detection of adversarial samples}
\textbf{Settings.} We use the same settings as in the ablation studies on vision to detect adversarial samples on the CIFAR-10 dataset. 
Adversarial samples are generated by using the same set (and same settings) of attacks used by~\citet{mahalanobis}, i.e. FGSM~\cite{fgsm}, BIM~\cite{bim}, DeepFool (DF)~\cite{deepfool} and CW~\cite{cw}. These attacks are generated against two classifiers with different architectures (ResNet and DenseNet) from~\citet{mahalanobis}'s paper, trained on CIFAR-10. We use the same AVT model as in the ablation studies for detecting adversarial samples against both architectures.

\textbf{\\ Comparison with SOTA and results.} We compare the performance of iDECODe with supervised detectors such as LID~\cite{lid}, Mahala~\cite{mahalanobis} and a detector based on combining of kernel density estimation~\cite{kd_pu} and predictive uncertainty (KD+PU). These detectors are trained on adversarial samples generated by FGSM attack. Unsupervised detectors such as Odds-testing~\cite{odds_testing} and AE-layers~\cite{ae_layers} are also considered. The details about these detectors are given in 
Appendix C.3.
Table~\ref{table:adv_table1} shows that iDECODe with $|\mathcal V(x)|=5$ achieves SOTA performance on DF and CW attacks against both architectures. Although it could not achieve SOTA results on FGSM and BIM attacks in comparison to the supervised detectors (Mahala and LID saw adversarial data generated by FGSM during its training and BIM is an iterative version of FGSM), it performs consistently well on these two attacks against both architectures. The results for supervised and unsupervised detectors are taken from ~\citet{mahalanobis} and~\citet{ae_layers}, respectively. 

\section{Conclusion}
\label{sec:conc}
We propose a new OOD detection method iDECODe, which leverages iD equivariance of a learned model with respect to a set of transforms. iDECODe uses a novel base NCM and a new aggregation method in ICAD framework, which guarantees a bounded FDR. We illustrate the efficacy of iDECODE on OOD detection for image and audio datasets. We also show that iDECODe performs consistently well on adversarial detection across different architectures.

\section*{Acknowledgement}
The authors acknowledge support in part by 
ARO W911NF-17-2-0196, AFRL/DARPA FA8750-18-C-0090, 
and NSF \#1740079. The views, opinions and/or findings expressed are those of the author(s) and should not be interpreted as representing the official views or policies of the U.S. Government.

\bibliography{main}

\appendix
\appendix
\section{A. Proof of Theorem 1}
\label{apdx:justification_F_proof}

Recall that data points in the proper training set and the calibration set are IID according to the probability
distribution $D$. If $x \sim D$, then the vectors
$\mathcal V(x), \mathcal V(x_{m+1}),\ldots, \mathcal V(x_l),$ are also IID conditioned on the proper training set and the set of transforms $\{g_{xi}, 1\leq i \leq n\} \cup \{g_{ji}: m+1\leq j \leq l, 1 \leq i \leq n\}$. Indeed, the $n+1$-dimensional vectors 
$(x,g_{x1},\ldots, g_{xn})$, and $(x_j,g_{j1},\ldots, g_{jn})$ where $j = m+1,\ldots,l$ are IID from the product distribution $D\times Q_G^n$. The vectors $\mathcal V$ are constructed by applying the same function $\mathcal A$ from 
Equation 4
to these vectors. Moreover $\mathcal V$ only depends on the proper training set and the set of transforms $\{g_{xi}, 1\leq i \leq n\} \cup \{g_{ji}: m+1\leq j \leq l, 1 \leq i \leq n\}$. Hence, conditioning on these, the vectors $\mathcal V(x)$ and $\mathcal V(x_j)$ are IID. 

Similarly, the $k+1$ random variables $F(\mathcal V(x))$, $F(\mathcal V(x_j))$, $j=m+1,\ldots,l$ (recall $k = l-m$) are IID conditioned on the proper training set and the set of transforms $\{g_{xi}, 1\leq i \leq n\} \cup \{g_{ji}: m+1\leq j \leq l, 1 \leq i \leq n\}$. 
Under our assumption, we thus have $k+1$ IID random variables, each with a continuous density. 

Therefore, the $p$-value from 
Equation 5 
is uniformly distributed in $\{1/(k+1), 2/(k+1),...,1\}$. This is essentially the same claim that is at the core of the validity of conformal prediction, see e.g., Proposition 2.4 in \cite{icp}, Theorem 1.2 in \citep{cp}. Here we sketch the argument making a connection to order statistics.
Let us denote the random variables by $R_i$, $i=1,\ldots,k+1$, namely $R_1 = F(\mathcal V(x))$ and $R_a = F(\mathcal V(x_{a-1+m}))$ for $a=2,\ldots k+1$. Then, the claim is that the number of indices $j\ge 1$ such that $R_1 \le R_j$ is uniformly distributed over $1,\ldots,k+1$. Now, this quantity is precisely the rank of $R_1$ among $R_1,R_2,\ldots,R_{k+1}$ minus unity. Therefore, from classical results on rank statistics, it follows that this random variable is uniformly distributed on $1,\ldots,k+1$, see e.g.,~\cite{lehmann1975nonparametrics,lehmann2006testing}.

\section{B. Proof of Corollary 1}
\label{apdx:bound_proof}

If the $p$-value for a given data point $x$ from Equation
6 
is less than $\epsilon \in (0, 1)$, then iDECODe outputs OOD as the answer. Since the $p$-value is distributed uniformly over $\{1/(k+1),\ldots, 1\}$, the probability of this event equals
$$\sum_{1\le j \le (k+1)\ep} 1/(k+1) = \lfloor (k+1)\ep \rfloor/(k+1) \le\ep.$$
This error is thus upper bounded by $\epsilon$. 

\section{C. Experimental details and additional results}
\label{apdx:exp}


\textbf{Evaluation metrics.} We call iD data points positives and OOD points negatives. We abbreivate by TP, FN, TN, and FP the notions of true positive, false negative, true negative and false positive. We use True Negative Rate (TNR) at 90\% True Positive Rate (TPR), and the Area under Receiver Operating characteristic curve (AUROC) for evaluation. 
\\
The first metric is TNR = TN/(TN+FP), when TPR = TP/(TP+FN) is 90\%. It indicates the percentage of OOD datapoints detected correctly when 90\% of the iD samples are detected correctly. AUROC plots TPR against the false positive rate FP/(FP+TN) by varying detection thresholds. Intuitively, AUROC indicates the probability of the $p$-value for OOD samples being lower than the $p$-value of the iD samples. Higher TNR and AUROC scores indicate a good performance of the OOD detector.

\subsection{C.1. Vision}
\label{apdx:vision}
\subsubsection{C.1.1. Details about ablation studies}
\label{apdx:ablation}

\textbf{\\ Datasets.} The CIFAR-10 dataset~\citep{cifar} consists of 60000 images for 10 classes, with 6000 images per class. The 10 classes in CIFAR-10 are airplane, car, bird, cat, deer, dog, frog, horse, ship, and trucks. There are 50000 training images (5000 per class) and 10000 test images (1000 per class). We randomly split the 50,000 training images into 45,000 proper training datapoints and 5000 calibration points. Following the experimental conventions for OOD detection, ~\citep[see e.g.,][]{mahalanobis, aux}, we use SVHN~\citep{svhn}, LSUN~\citep{lsun}, ImageNet~\citep{imagenet}, CIFAR100~\citep{cifar}, and Places365~\citep{places} as the OOD datasets for CIFAR-10. Street View House Numbers (SVHN) is a dataset containing numbers taken from the Google Street View. Places365 is a scene recognition dataset. LSUN is another street recognition dataset but ten times larger in size than Places365. Imagenet is an object recognition dataset. CIFAR100 is the 100-class version of CIFAR-10. The classes in CIFAR-10 and CIFAR100 do not overlap, so CIFAR100 is an OOD dataset for CIFAR-10.

\textbf{\\ Model details.} We use ResNet34~\citep{resnet} architecture for the AVT model trained on the CIFAR-10 dataset. 
Following the architecture of the AVT model on CIFAR-10 from~\citet{qi2019}, the encoder block consists of a convolutional layer followed by batchnorm and is inserted before the last block of ResNet34. The last block is followed by the global average pool layer. The original and transformed images $x$, $g(x), g \in G$, are fed into this architecture. The average-pooled features of the original and the transformed images are concatenated and fed into the two layer fully connected decoder network. This is the same decoder architecture used by~\citet{qi2019}. The model is trained with the same loss function from~\citet{qi2019}'s AVT model on CIFAR-10, i.e. the mean square loss between the actual and the decoder-predicted values of the transformation matrix.\footnote{Our code for model training is build on top \url{https://github.com/maple-research-lab/AVT-pytorch/tree/master/cifar}.}

\subsubsection{C.1.2. Additional results on CIFAR-10.}
\begin{figure*}[t!]
    \centering
    \begin{subfigure}
        \centering
        \includegraphics[height=0.7in]{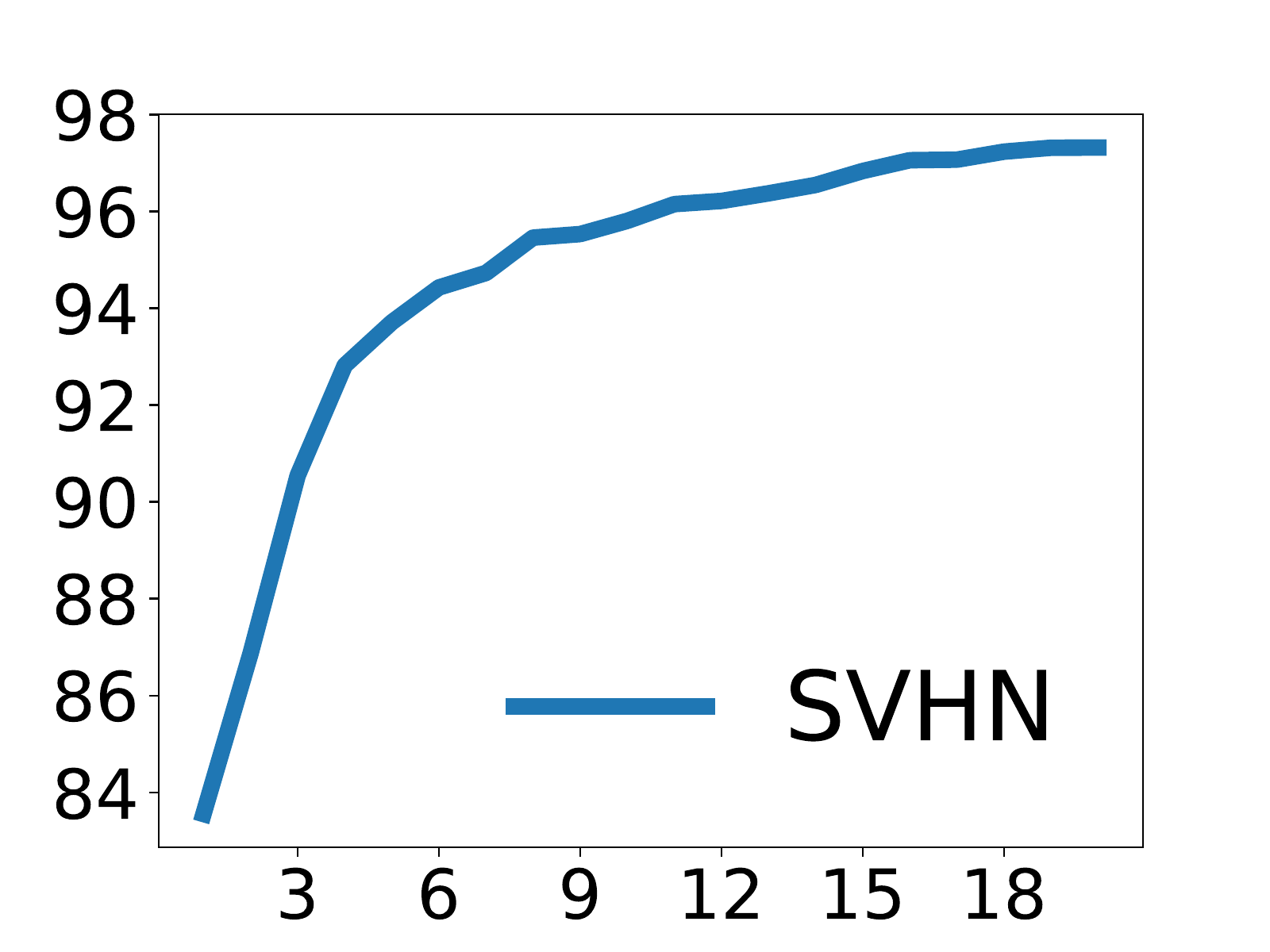}
    \end{subfigure}%
    ~ 
    \begin{subfigure}
        \centering
        \includegraphics[height=0.7in]{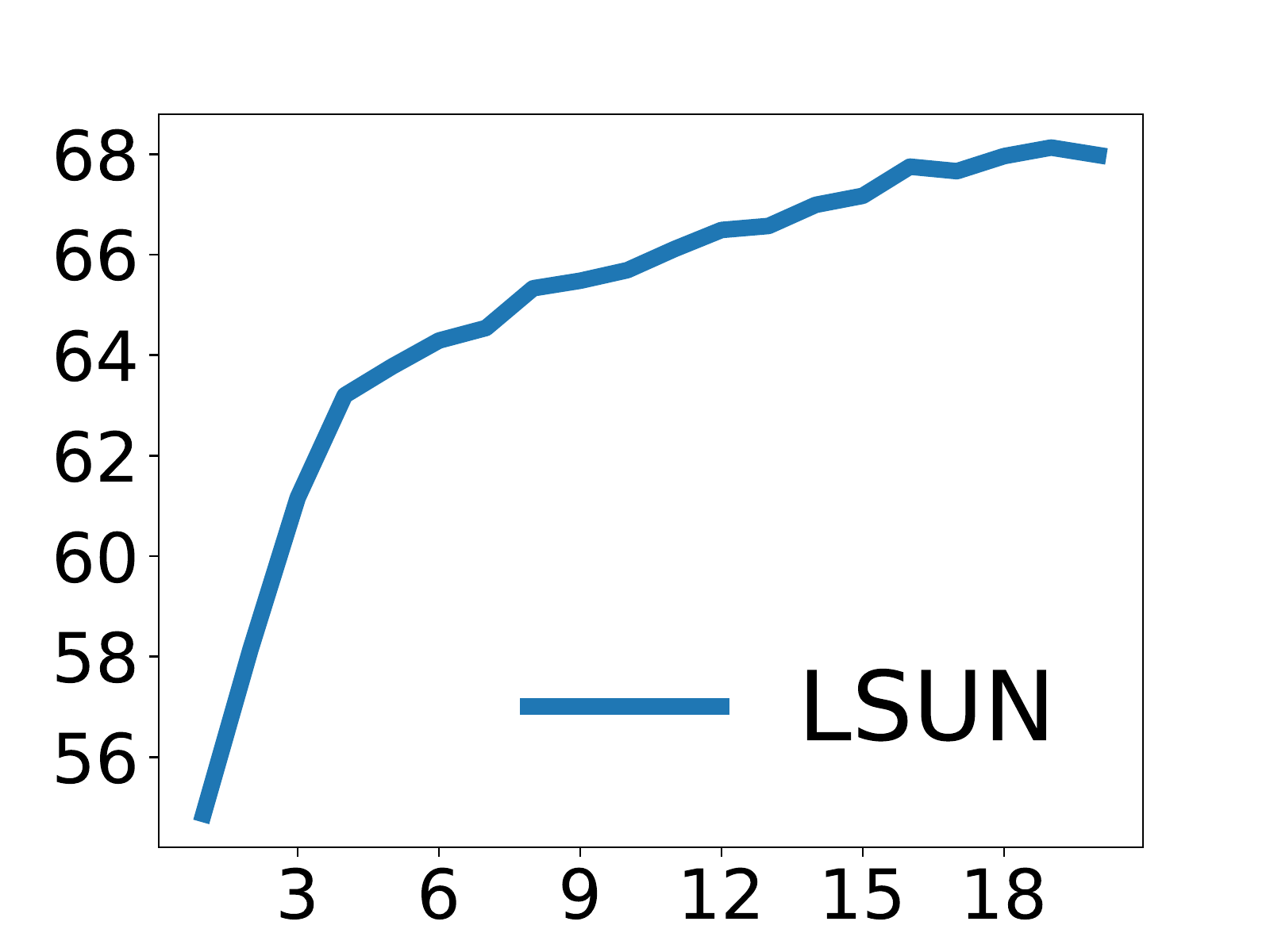}
    \end{subfigure}
    ~ 
    \begin{subfigure}
        \centering
        \includegraphics[height=0.7in]{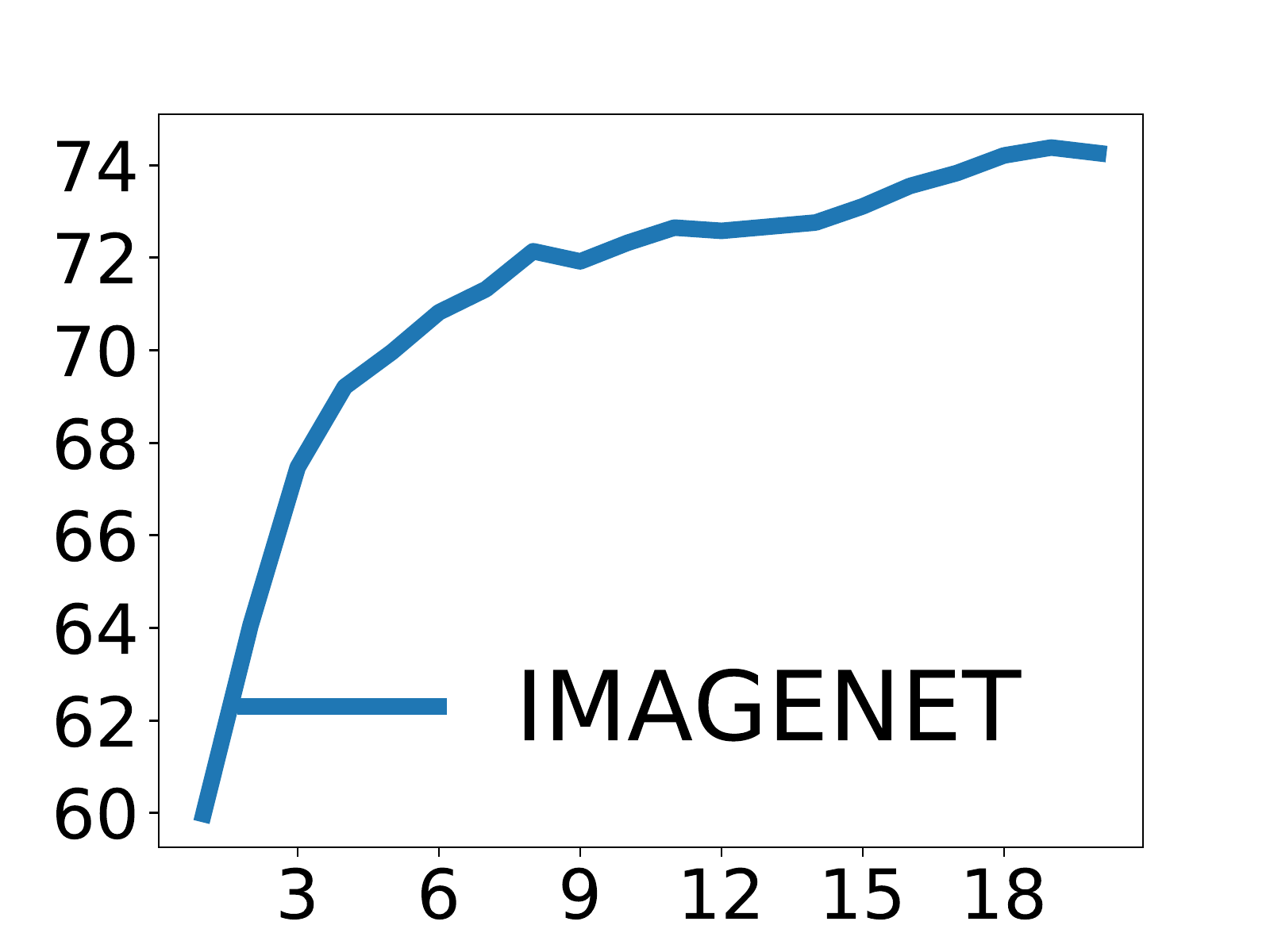}
    \end{subfigure}
    ~ 
    \begin{subfigure}
        \centering
        \includegraphics[height=0.7in]{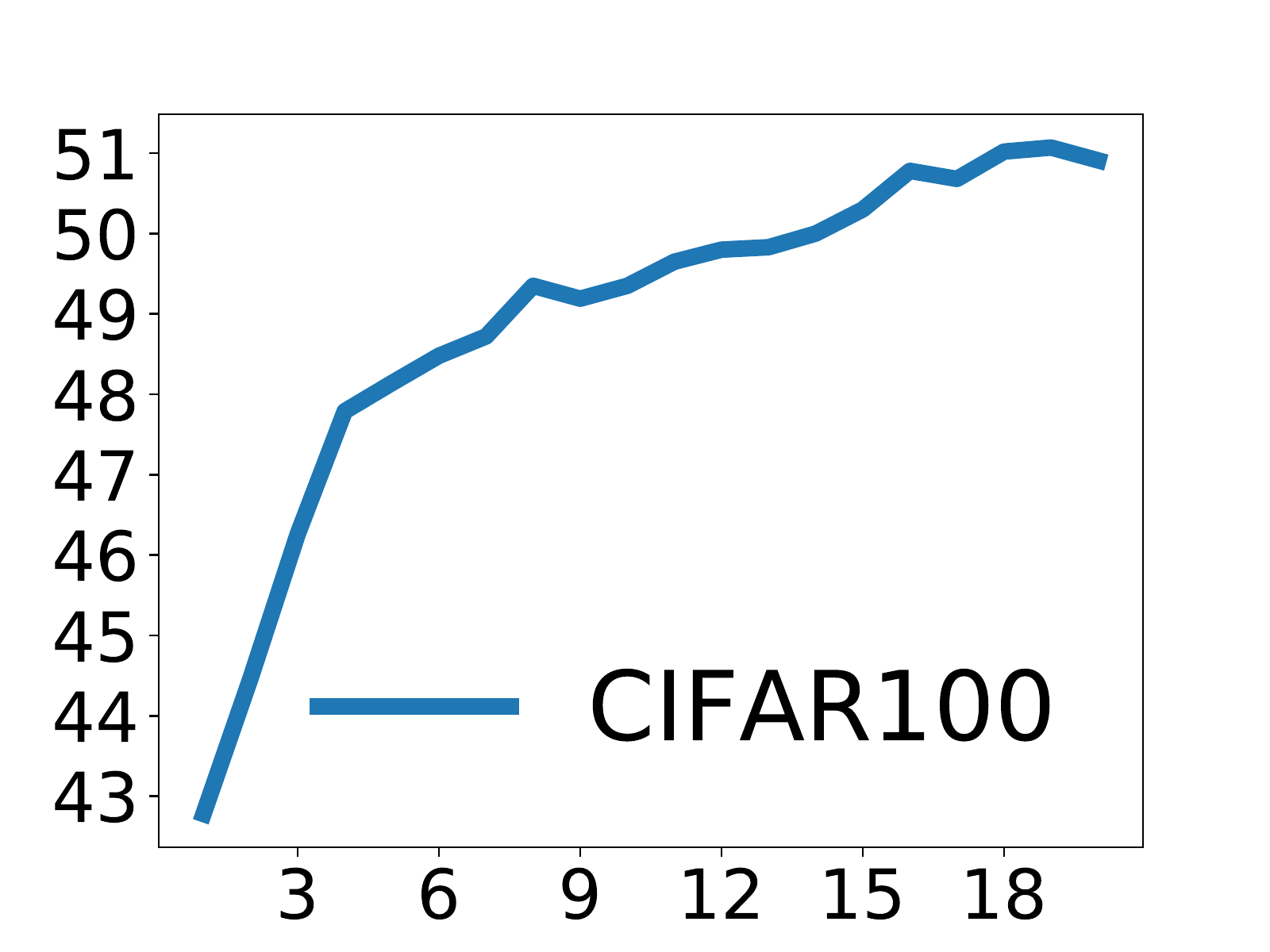}
    \end{subfigure}
    ~ 
    \begin{subfigure}
        \centering
        \includegraphics[height=0.7in]{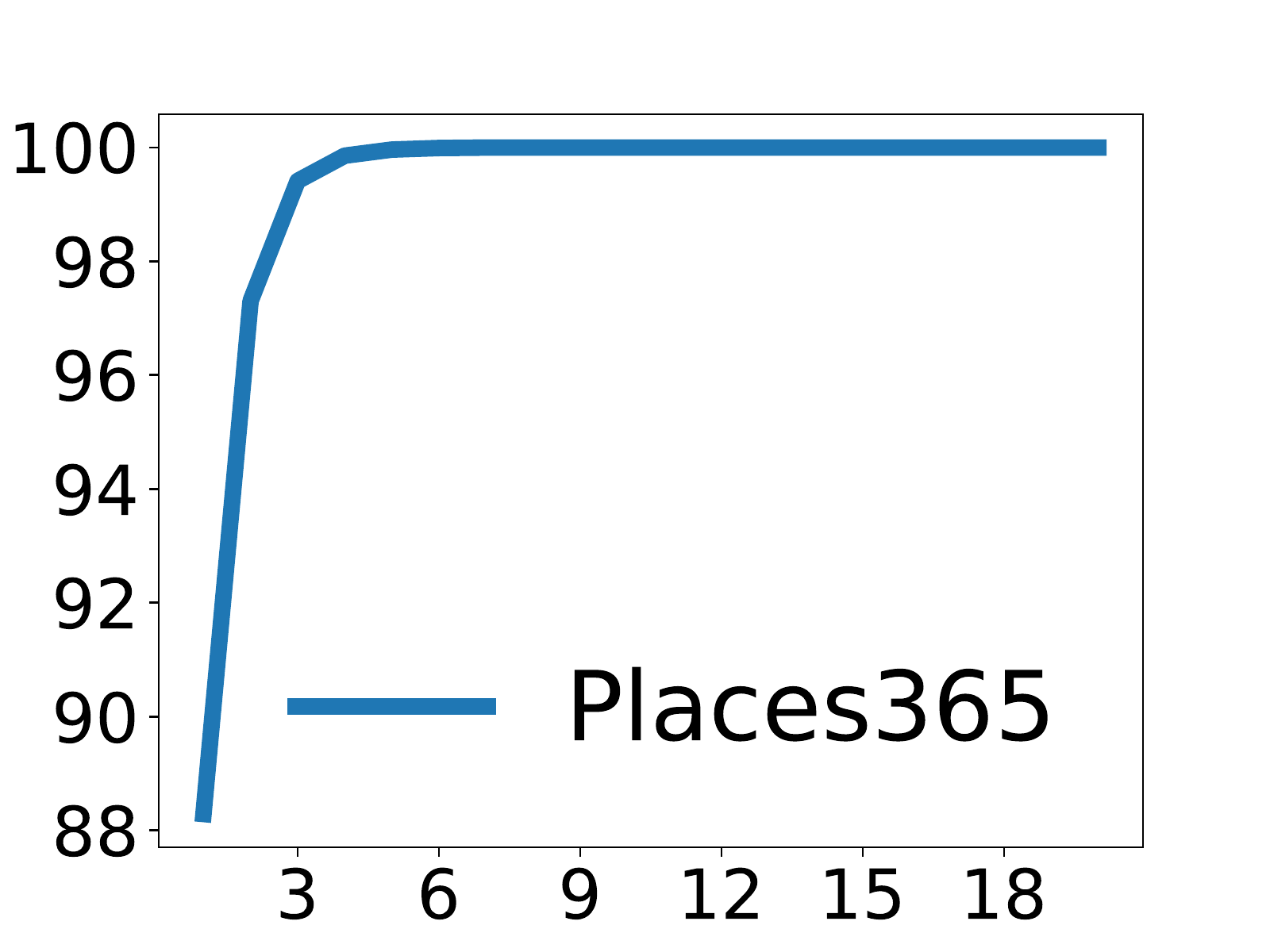}
    \end{subfigure}
    
    \caption{\footnotesize{TNR vs $|\mathcal V(x)|$ reported by iDECODe on CIFAR-10 as iD.}}
    \label{fig:tnr_n}
\end{figure*}

\begin{figure*}[!t]

    \centering
    \begin{subfigure}[]
        \centering
        \includegraphics[height=1.4in]{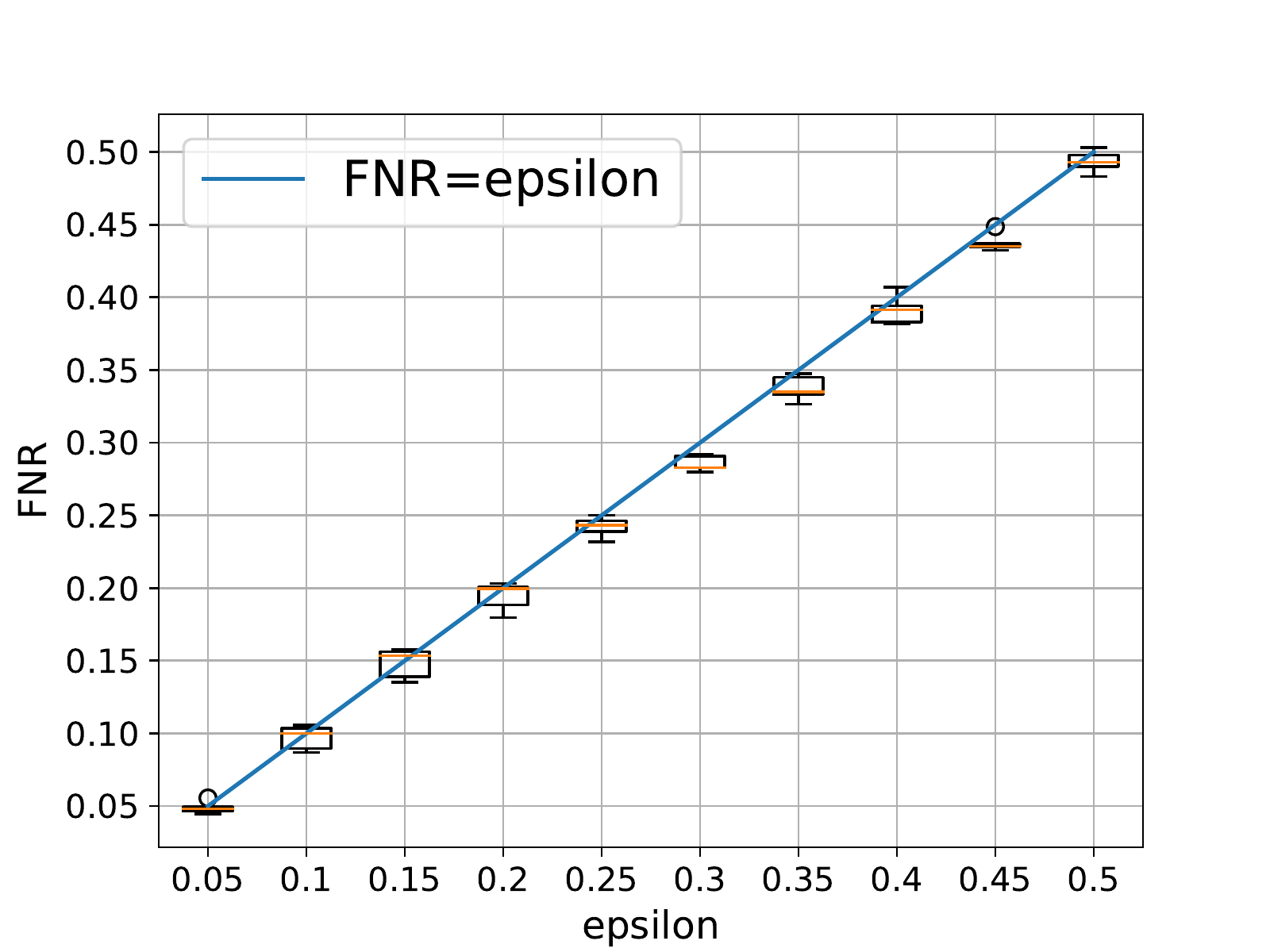}
    \end{subfigure}
    \quad
    \begin{subfigure}
        \centering
        \includegraphics[height=1.4in]{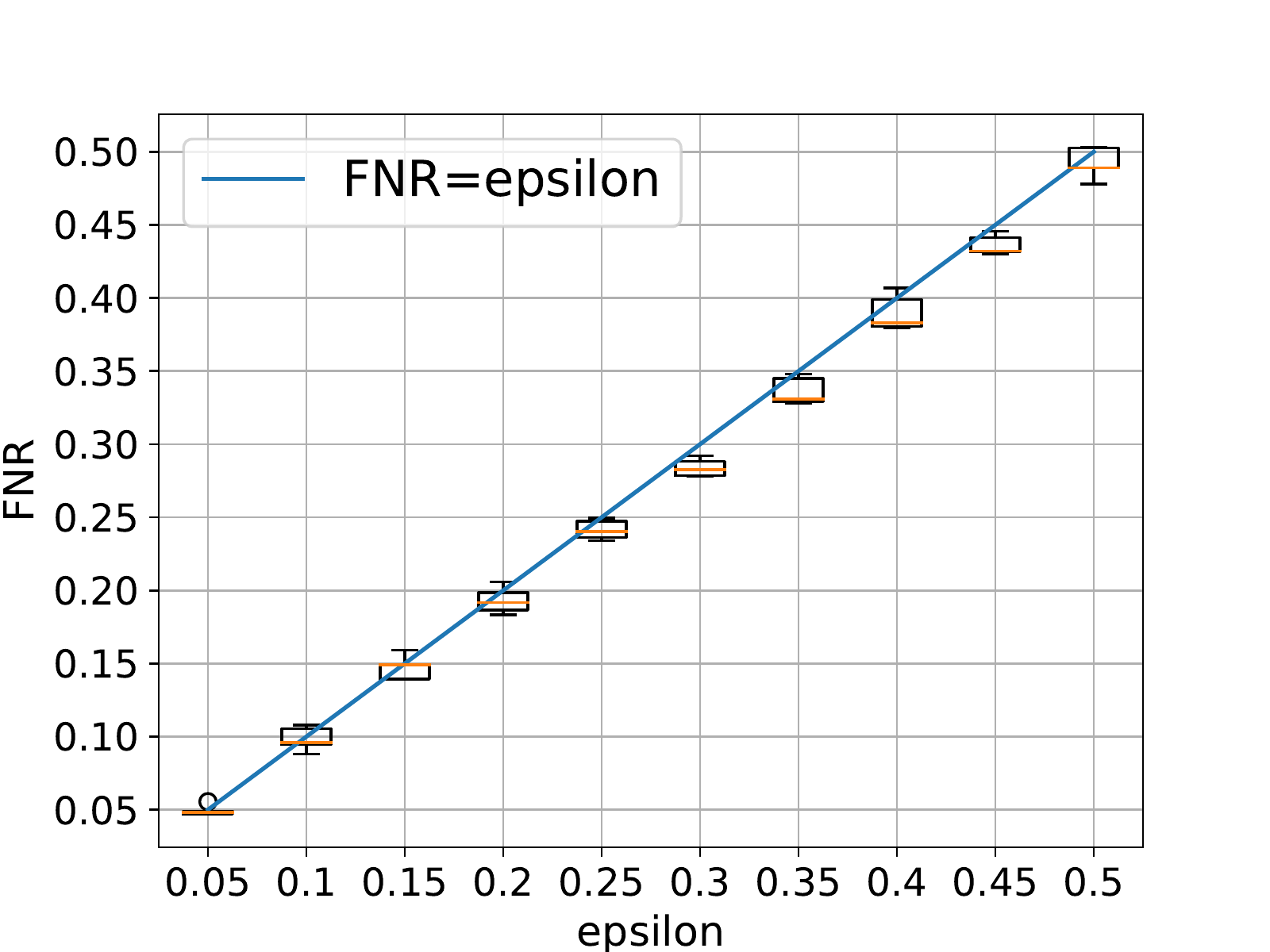}
    \end{subfigure}
    \quad
    \begin{subfigure}[]
        \centering
        \includegraphics[height=1.4in]{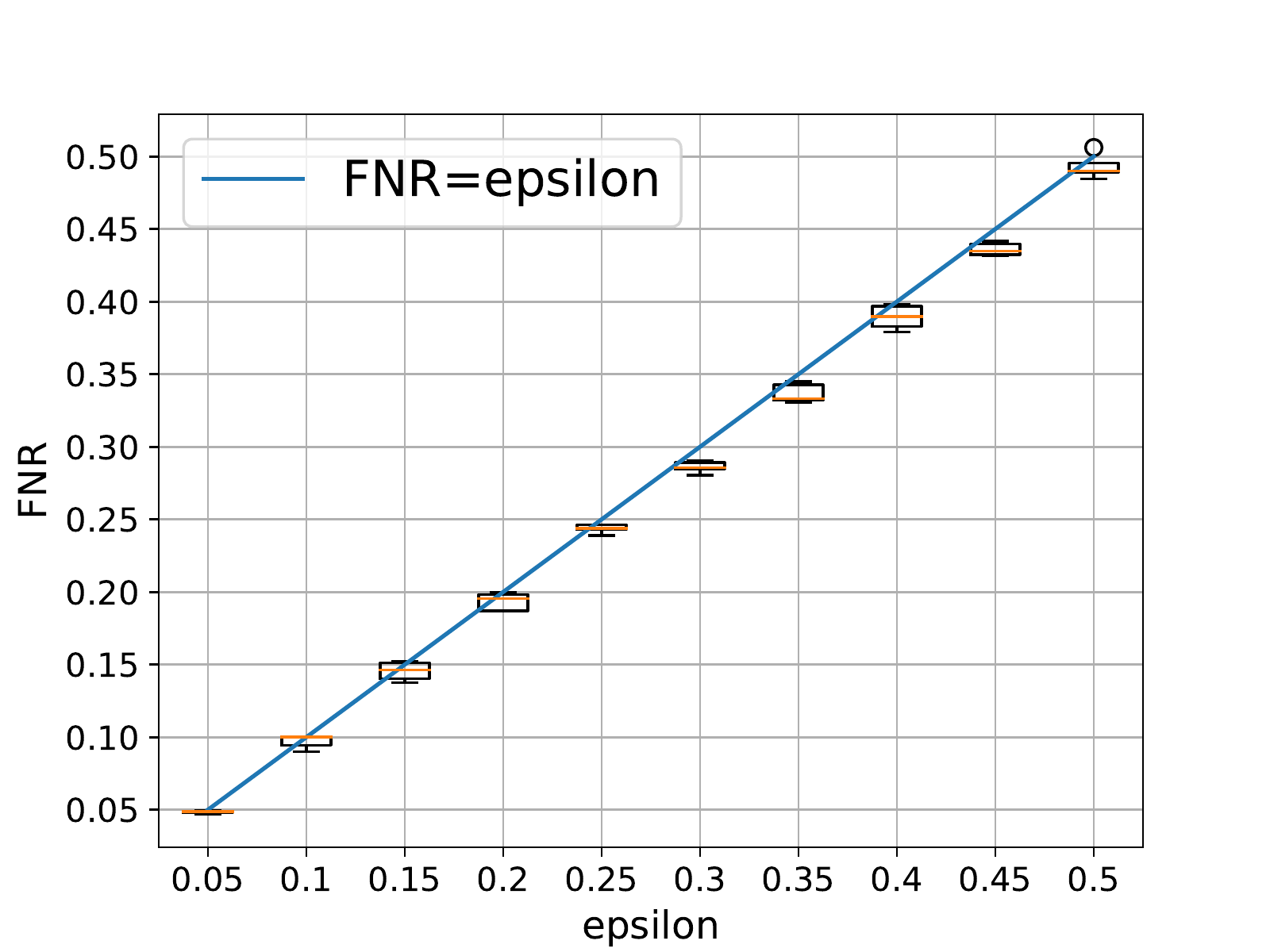}
    \end{subfigure}%
    \quad
    \begin{subfigure}
        \centering
        \includegraphics[height=1.4in]{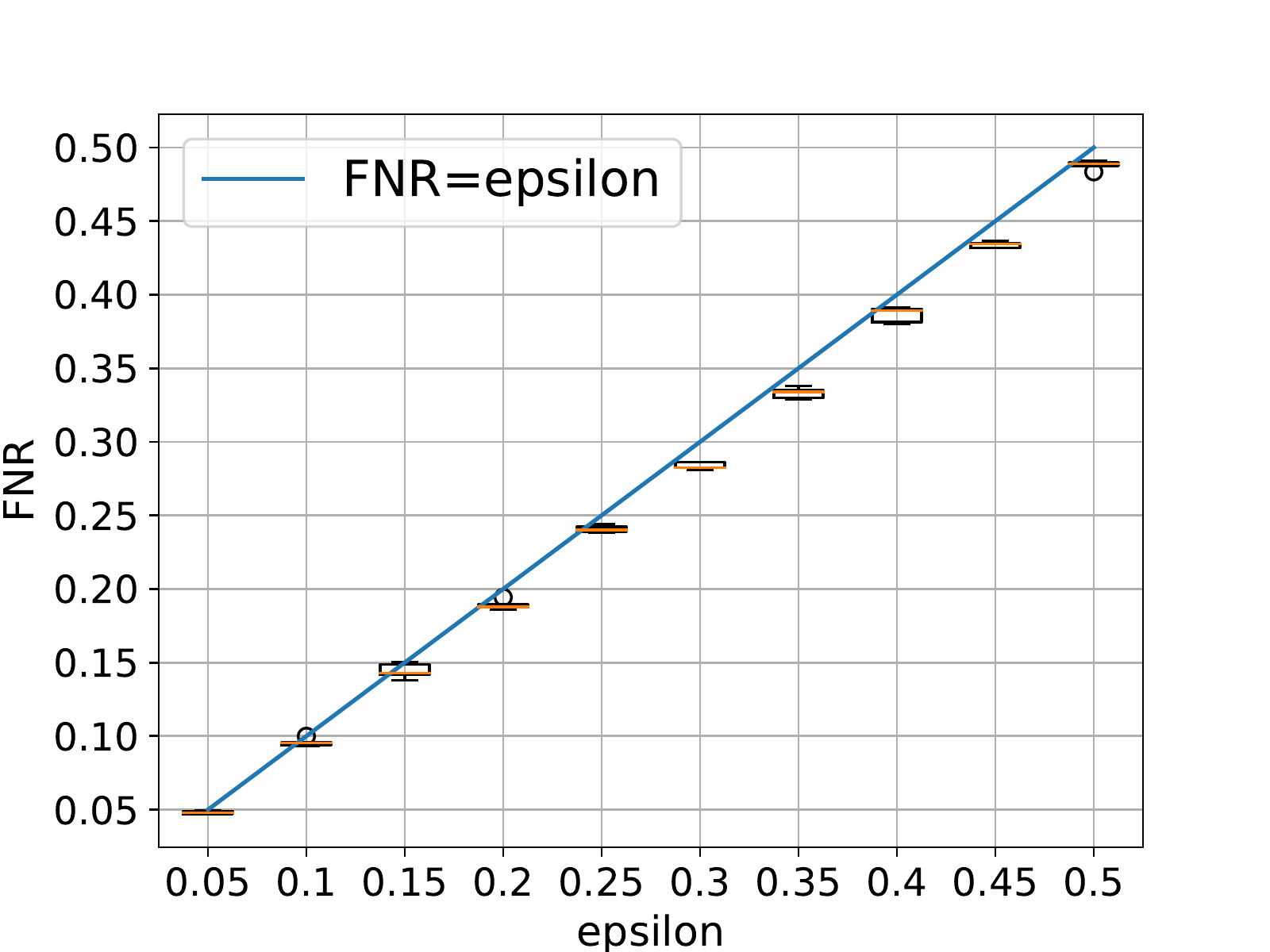}
    \end{subfigure}%
    \quad
    \begin{subfigure}[]
        \centering
        \includegraphics[height=1.4in]{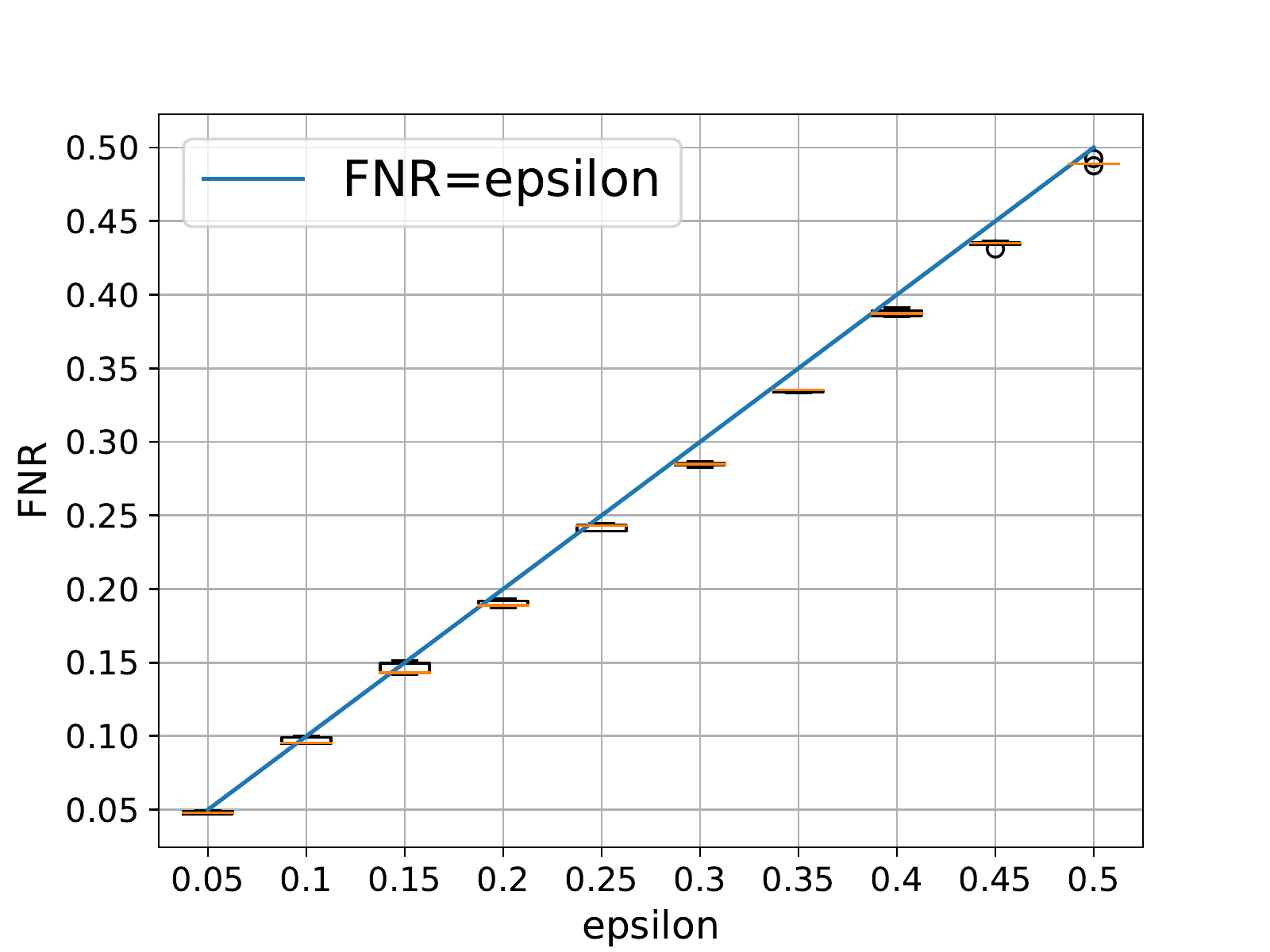}
    \end{subfigure}%
    \quad
    \begin{subfigure}
        \centering
        \includegraphics[height=1.4in]{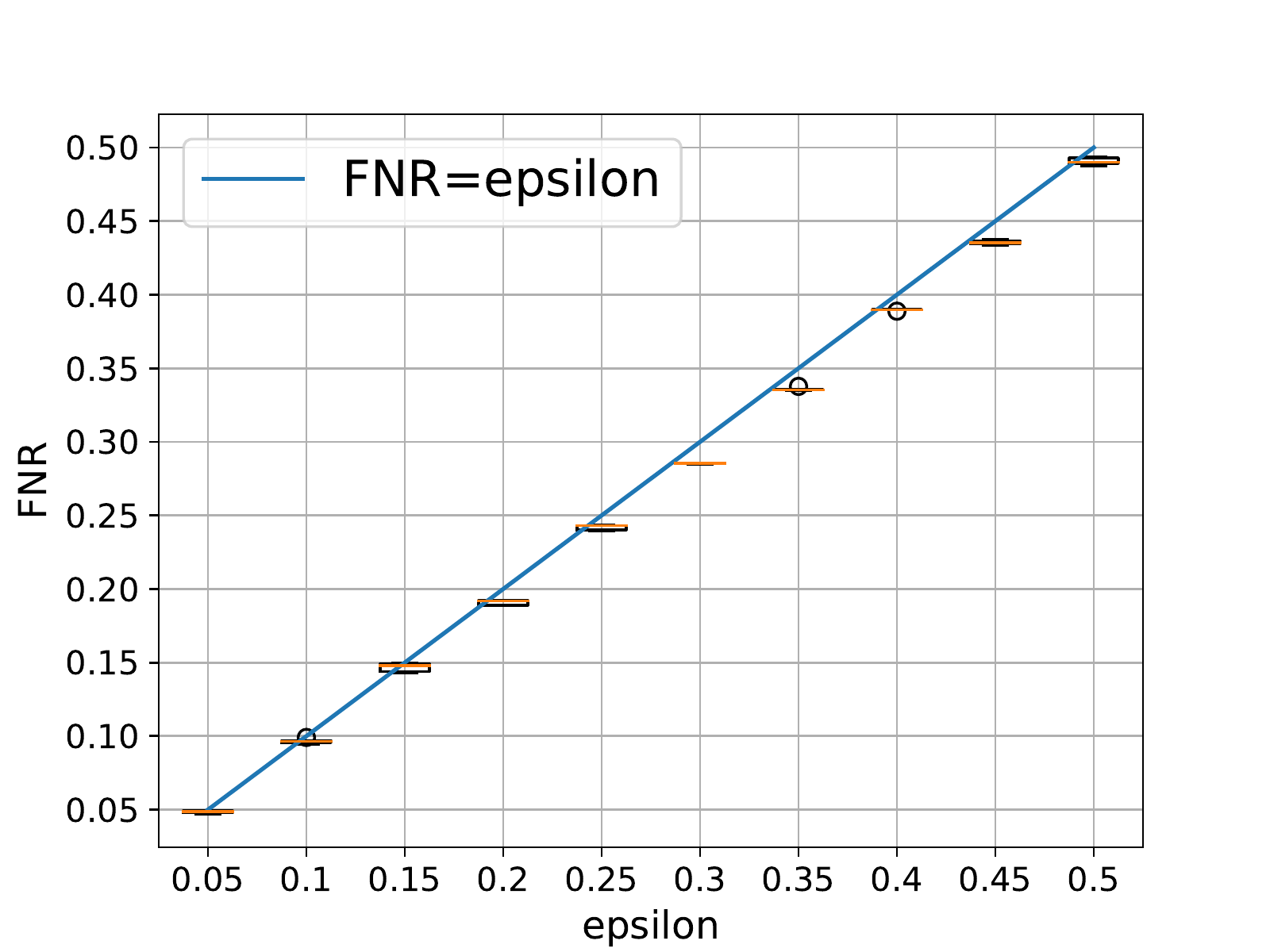}
    \end{subfigure}%
    \quad
    
    \caption{False detection rate of iDECODe is upper bounded by $\epsilon$ on average. From left to right, $|X_{\text{cal}}| = 1500, 2000, 2500, 3000, 3500, 4000$ on CIFAR-10 as iD.}
    
\label{more_box_plots} 
\end{figure*}

Figure~\ref{fig:tnr_n} shows that similar to AUROC, TNR (90\% TPR) also increases with the size of $\mathcal V(x) (|\mathcal V(x)| = 1, . . . , 20)$. 
Figure~\ref{more_box_plots} shows more results on the bounded false detection rate with different sizes of the sampled calibration set.

\subsubsection{C.1.3. SVHN as in-distribution for ablations study}
\label{apdx:svhn}
\textbf{\\ Settings.} We perform ablation studies with SVHN as iD by training the AVT model on SVHN. The model has the same architecture, hyperparameters and loss function as the model trained for ablation studies on CIFAR-10. We use the same set $G$ and the base NCM as in the ablations studies for CIFAR-10. For SVHN, we use CIFAR-10, LSUN, ImageNet, CIFAR100 and Places365 as OOD datasets.

\begin{figure*}[t!]
    \centering
    \begin{subfigure}
        \centering
        \includegraphics[height=0.7in]{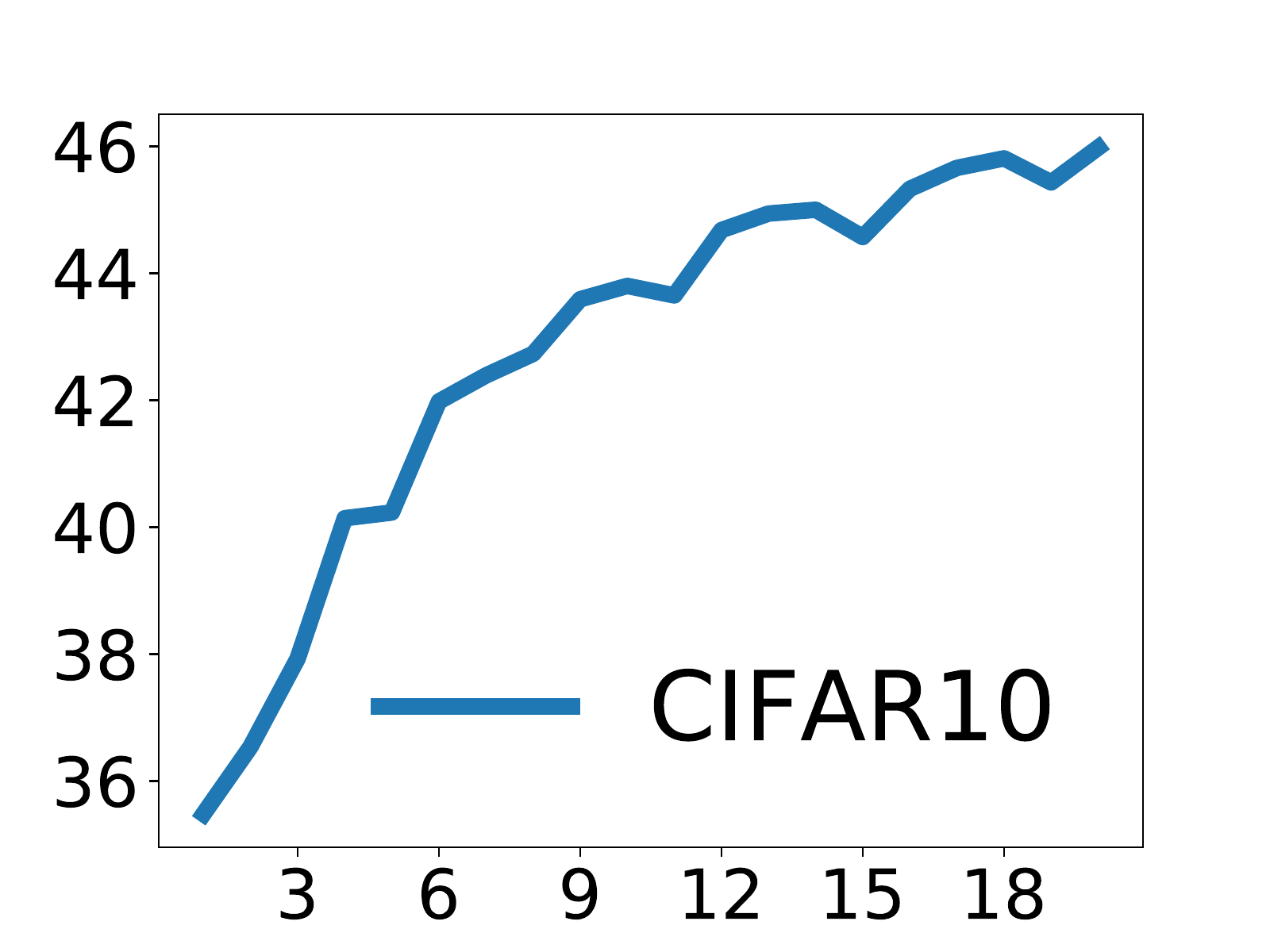}
    \end{subfigure}%
    ~ 
    \begin{subfigure}
        \centering
        \includegraphics[height=0.7in]{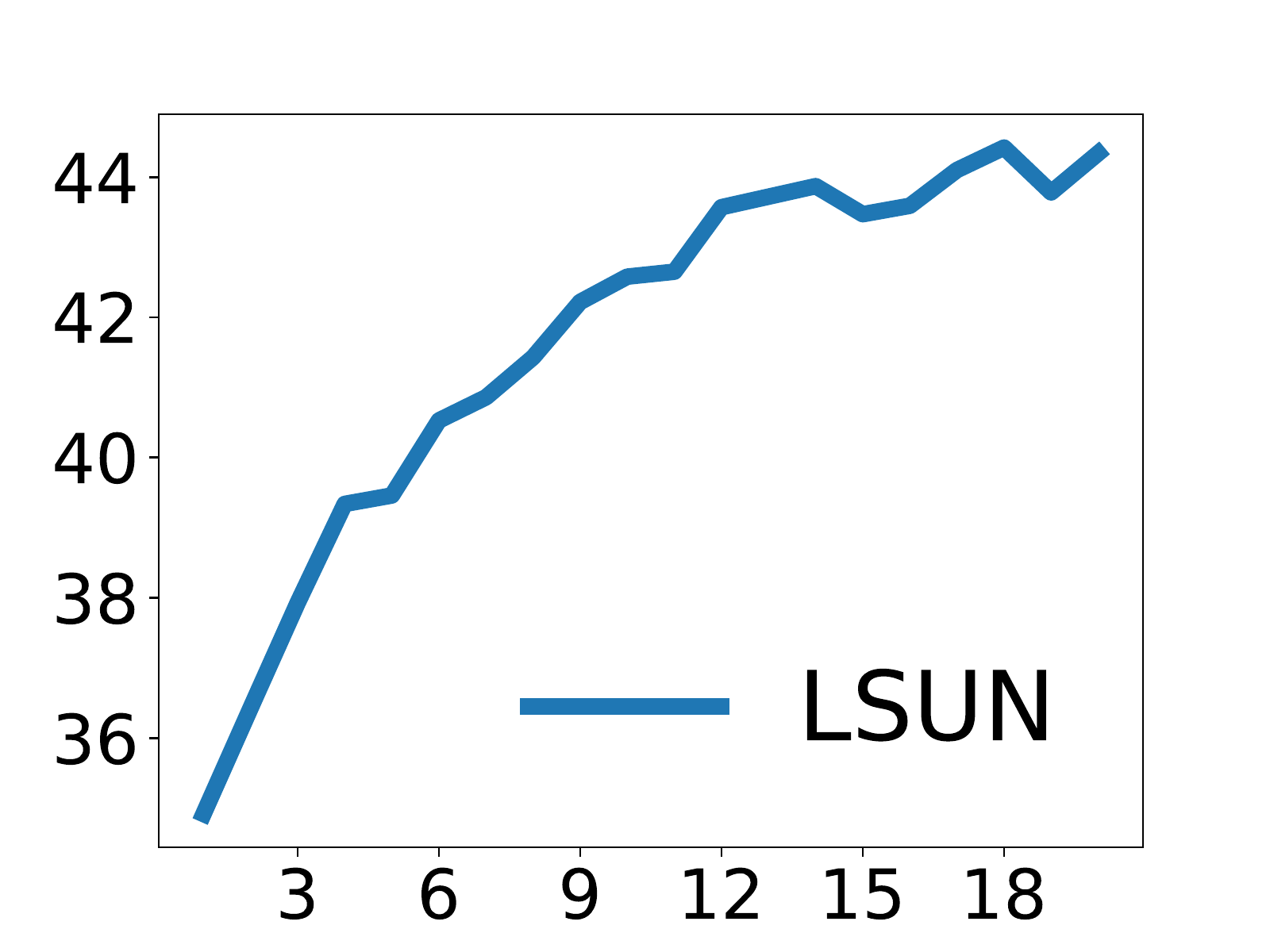}
    \end{subfigure}
    ~ 
    \begin{subfigure}
        \centering
        \includegraphics[height=0.7in]{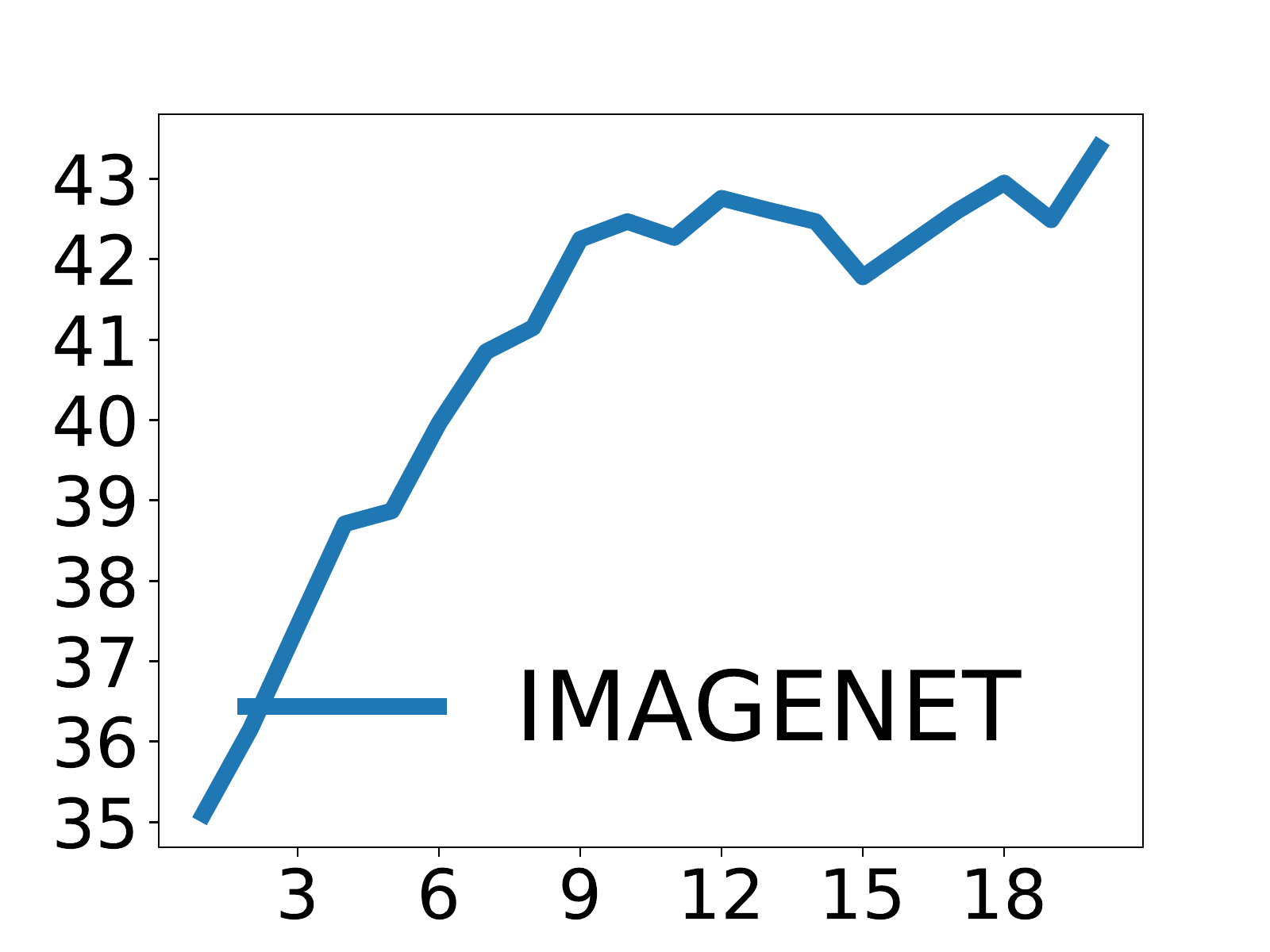}
    \end{subfigure}
    ~ 
    \begin{subfigure}
        \centering
        \includegraphics[height=0.7in]{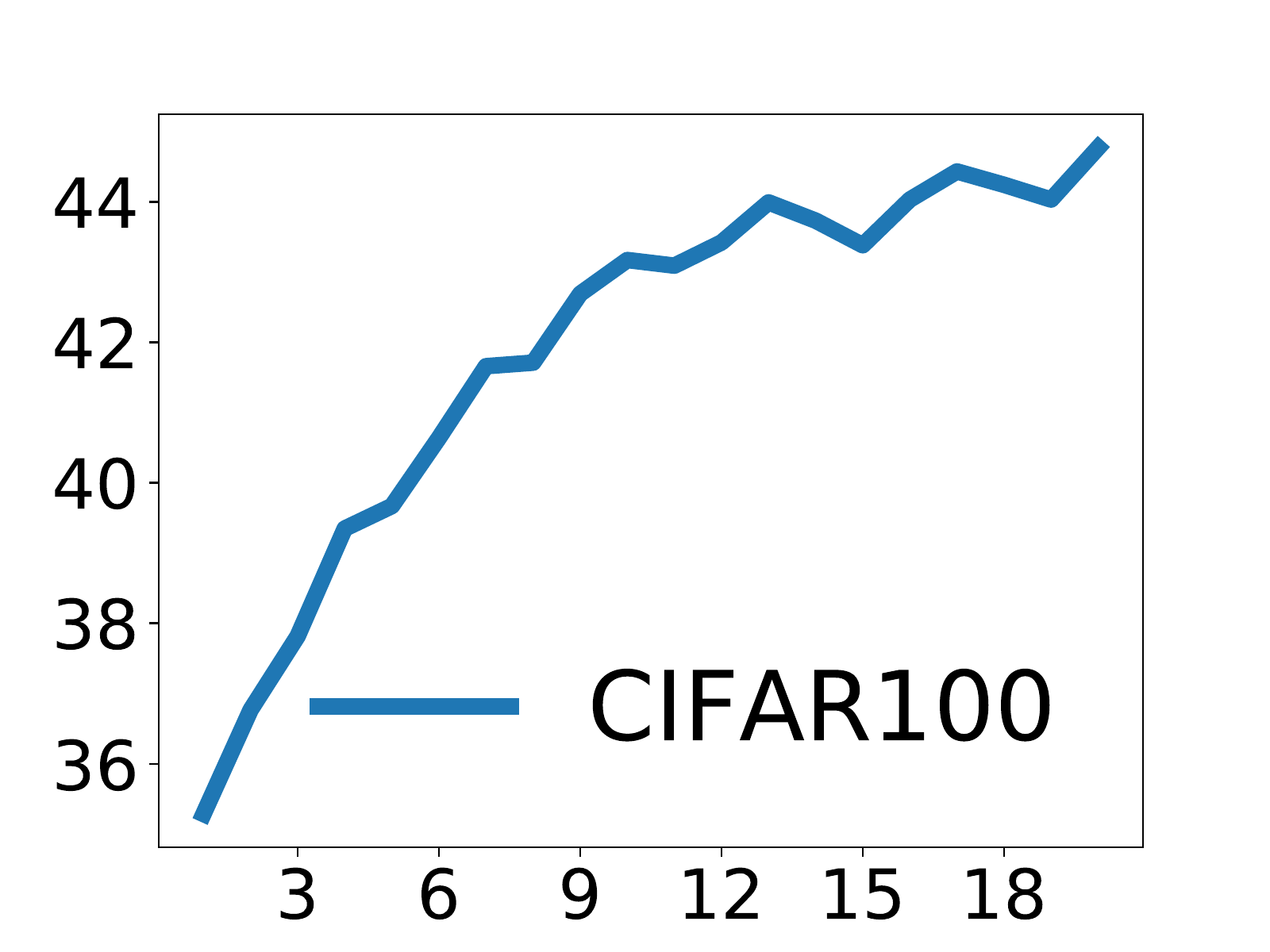}
    \end{subfigure}
    ~ 
    \begin{subfigure}
        \centering
        \includegraphics[height=0.7in]{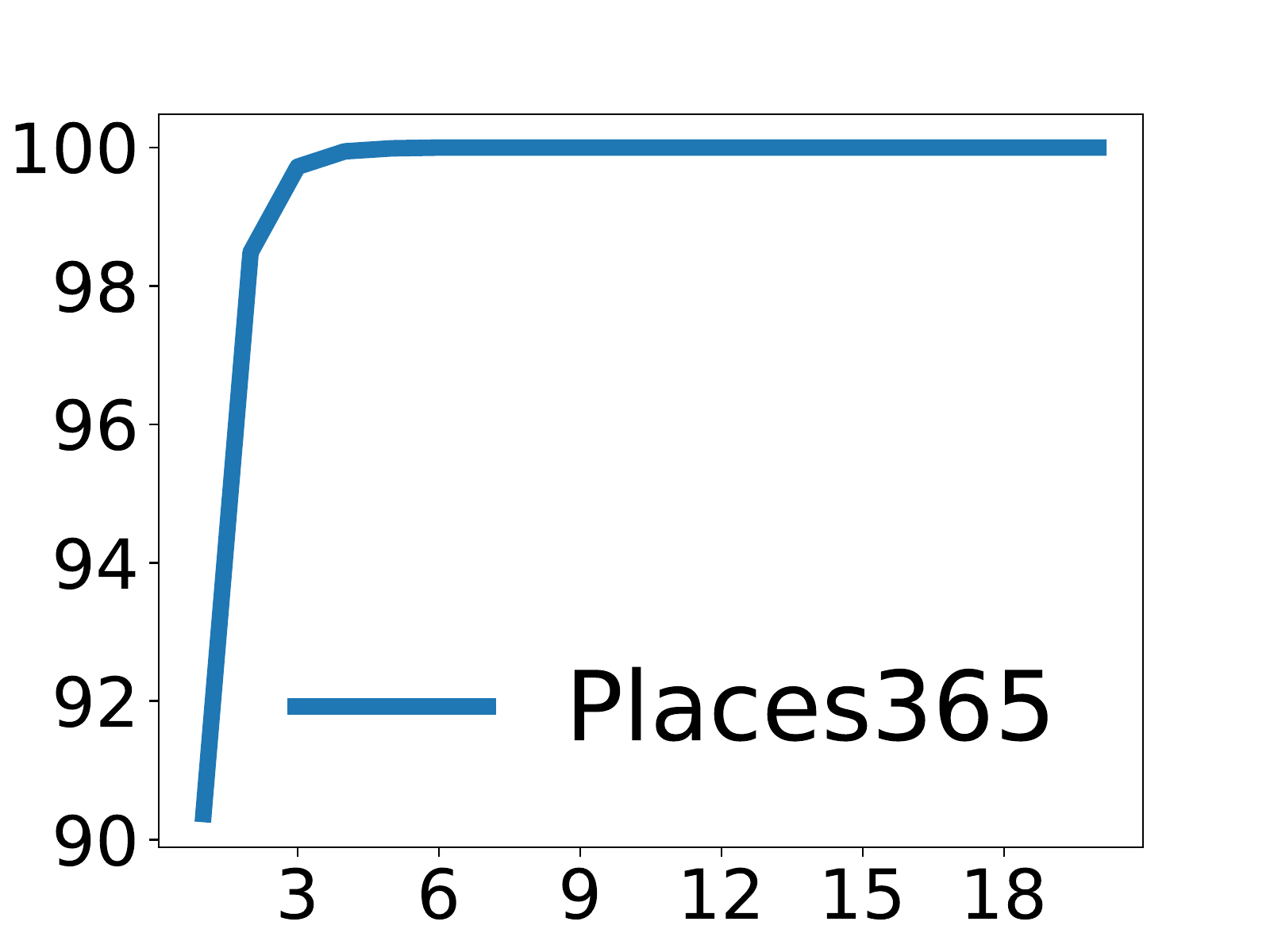}
    \end{subfigure}
    
    \centering
    \begin{subfigure}
        \centering
        \includegraphics[height=0.7in]{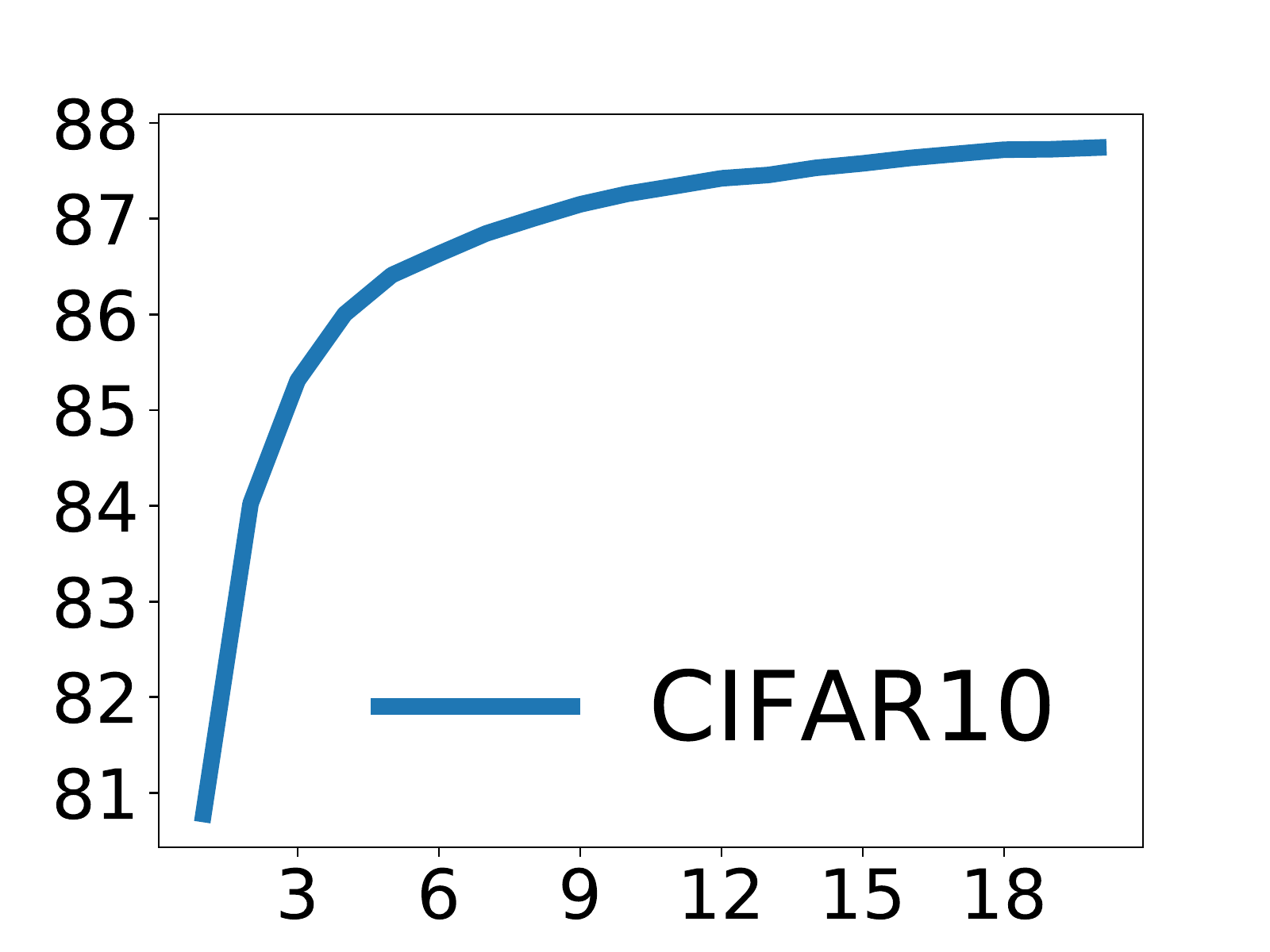}
    \end{subfigure}%
    ~ 
    \begin{subfigure}
        \centering
        \includegraphics[height=0.7in]{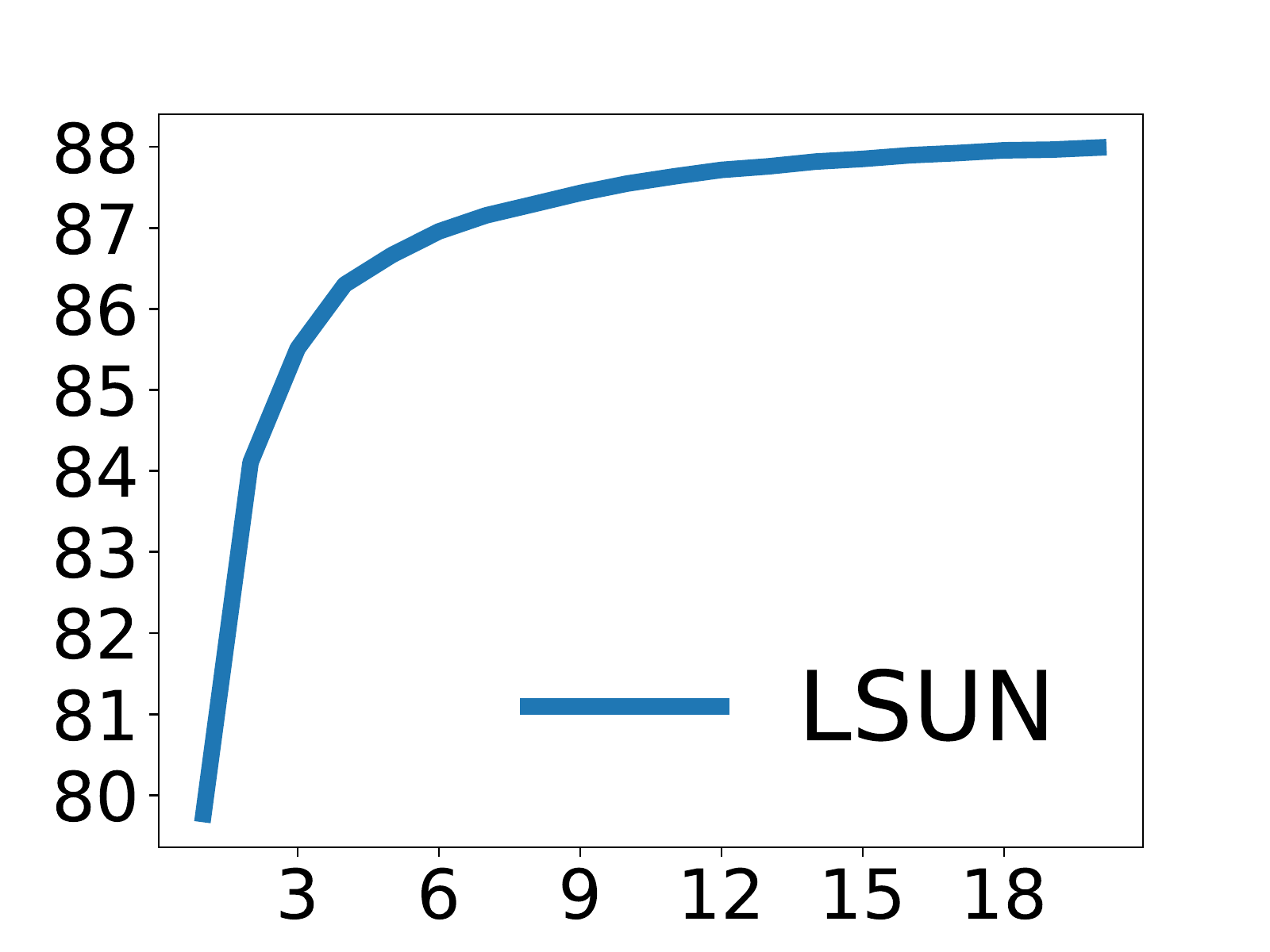}
    \end{subfigure}
    ~ 
    \begin{subfigure}
        \centering
        \includegraphics[height=0.7in]{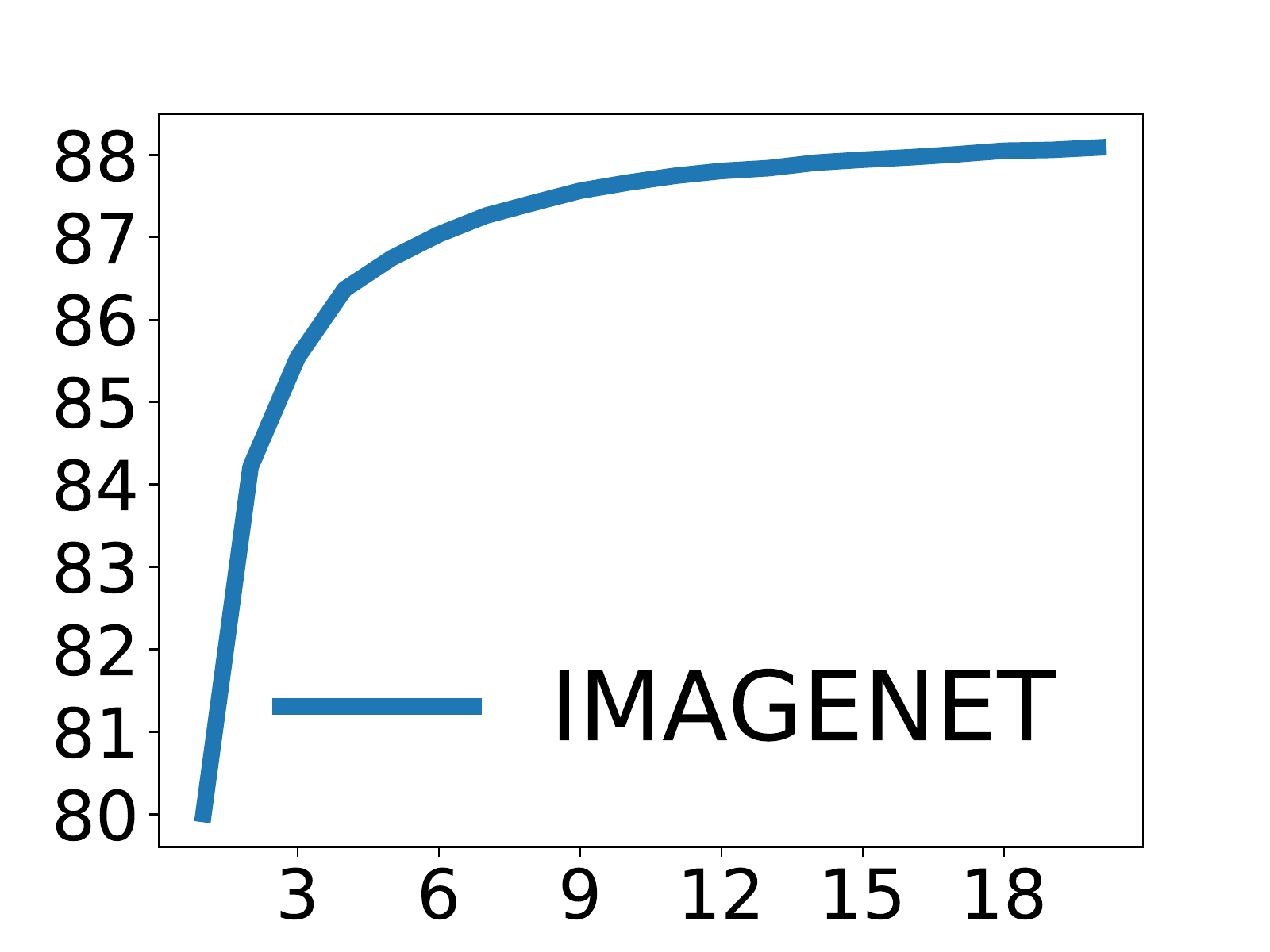}
    \end{subfigure}
    ~ 
    \begin{subfigure}
        \centering
        \includegraphics[height=0.7in]{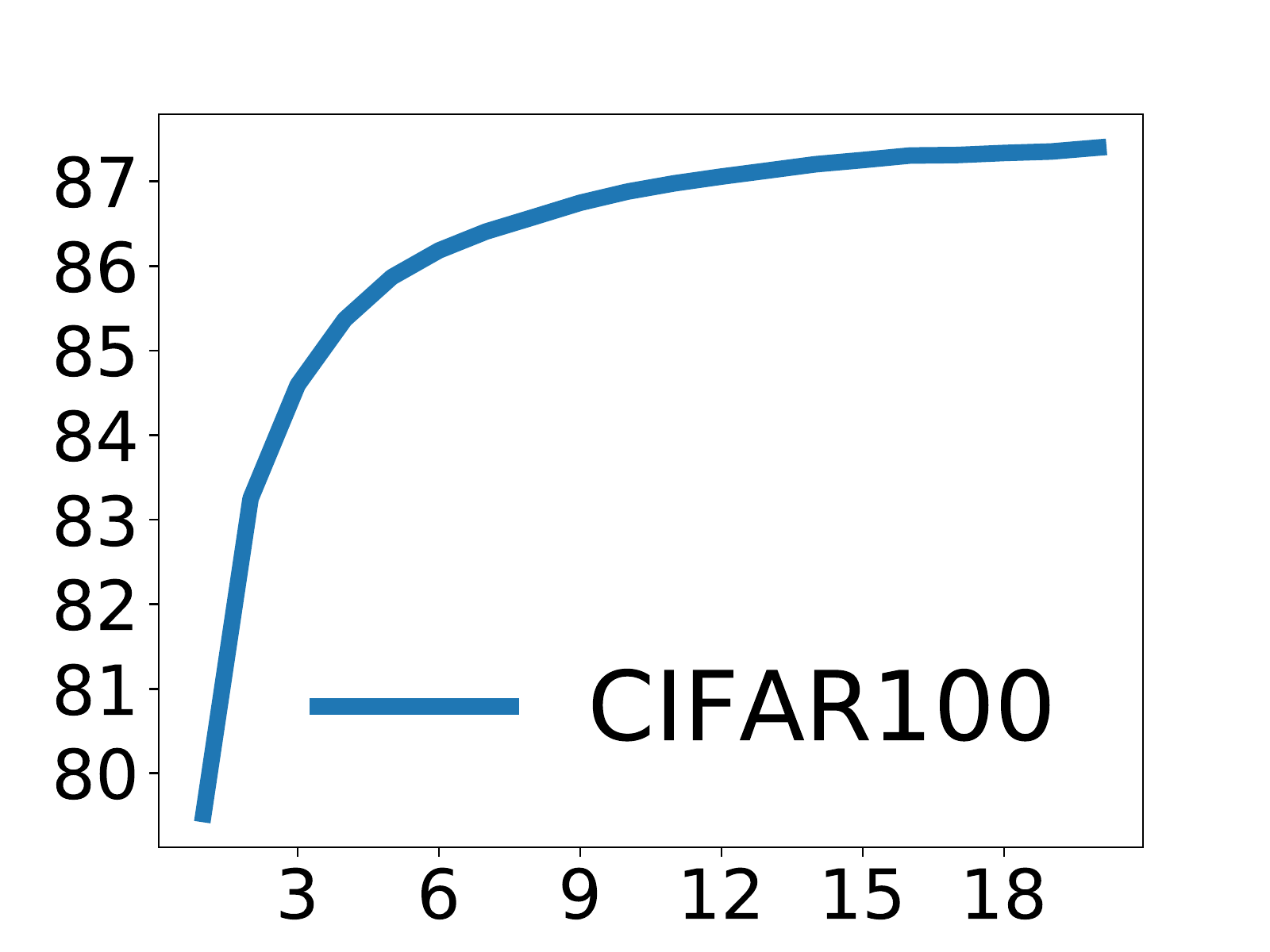}
    \end{subfigure}
    ~ 
    \begin{subfigure}
        \centering
        \includegraphics[height=0.7in]{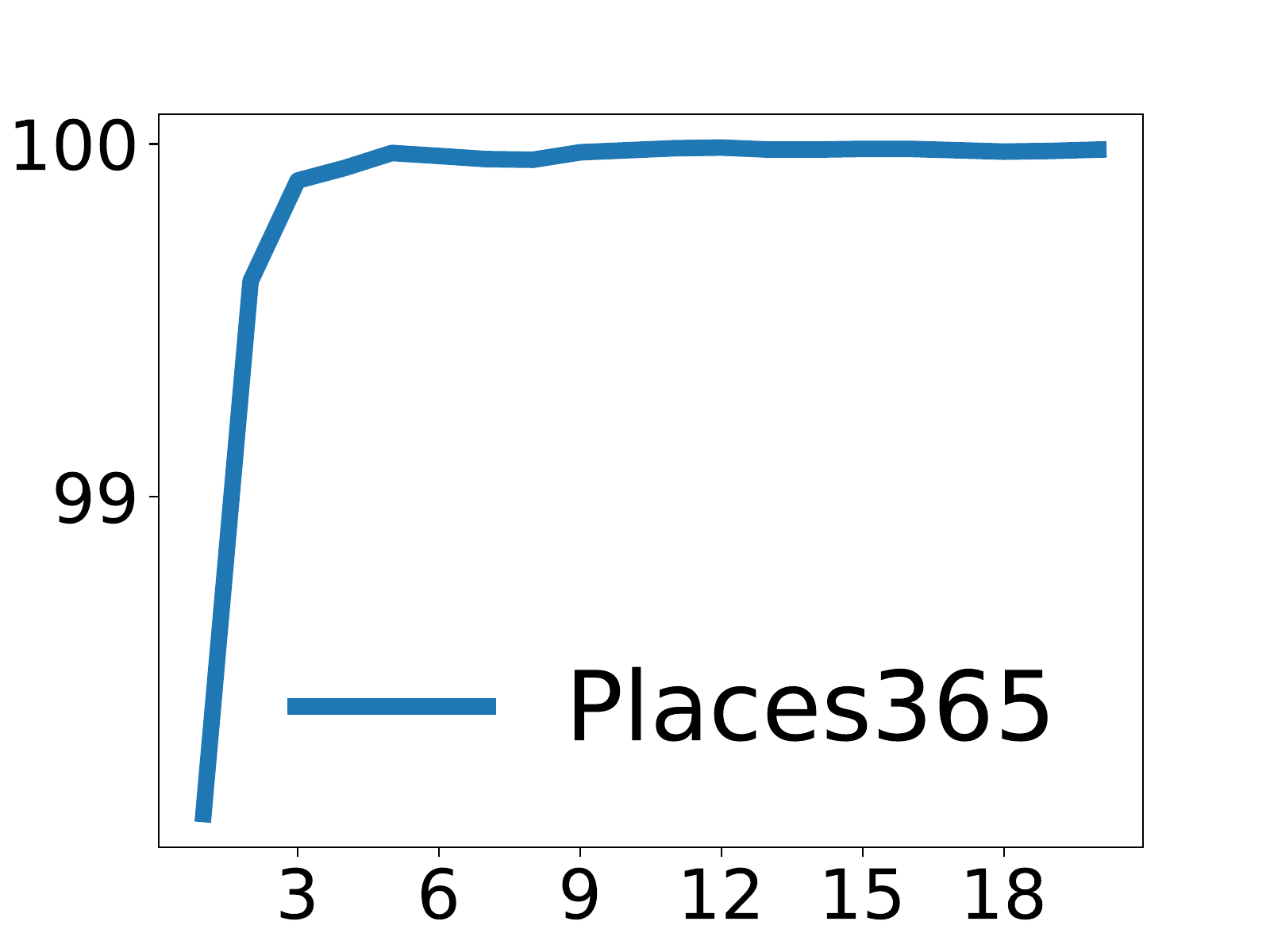}
    \end{subfigure}
    \caption{\footnotesize{TNR at 90\% TPR vs $|\mathcal V(x)|$ (top), AUROC vs $|\mathcal V(x)|$ (bottom) by iDECODe on SVHN as iD.}}
    \label{fig:svhn_tnr_auroc_n}
\end{figure*}

\textbf{\\ Results.} Table~\ref{tab:comp_base_icad_svhn} compares the performance of the base score method, ICAD and iDECODe with $|\mathcal V(x)|=5$ (Ours). Ours outperforms both the base score method and ICAD on all OOD datasets for both TNR at 90\% TPR and AUROC. The results are averaged over five runs of randomly sampled set of transforms. Figure~\ref{fig:svhn_tnr_auroc_n} shows that the performance of iDECODe improves with the increase in $|\mathcal V(x)|$ for both TNR at 90\% TPR and AUROC. These results are similar to the results obtained from ablation studies on CIFAR-10 as iD.

\begin{table*}
\caption {\label{tab:comp_base_icad_svhn}Comparison of base score method and ICAD with iDECODe on SVHN as iD.} 
\begin{center}
\begin{adjustbox}{width=1.6\columnwidth,center}
\begin{tabular}{c|ccc|ccc}
\hline
$D_{out}$  & \multicolumn{3}{c|}{TNR (90\% TPR)}  &   \multicolumn{3}{c}{AUROC}     \\ 
\cline{2-4} \cline{5-7}  
        & Base Score     & ICAD  & Ours                                                         
	&  Base Score     & ICAD   & Ours                                              \\ 
\hline
CIFAR-10  & 36.01 $\pm$ 0.52  & 35.87 $\pm$ 0.55 & \textbf{40.75$\pm$ 0.38} 
    & 80.74 $\pm$ 0.13 & 80.74 $\pm$ 0.13 & \textbf{86.35 $\pm$ 0.07}\\ 
LSUN &  35.66 $\pm$ 0.62  &  35.51 $\pm$ 0.64  & \textbf{39.99 $\pm$ 0.56}
    & 80.08 $\pm$ 0.27 & 80.08$\pm$ 0.27  & \textbf{86.72 $\pm$ 0.09}\\
ImageNet &  35.65 $\pm$ 0.40  &  35.50 $\pm$ 0.41 & \textbf{39.53 $\pm$ 0.51}
    & 80.21 $\pm$ 0.14  & 80.21 $\pm$ 0.14 & \textbf{86.78 $\pm$ 0.11}\\
CIFAR100 & 35.88 $\pm$ 0.43  & 35.74 $\pm$ 0.43   &   \textbf{40.41 $\pm$ 0.41}
    & 79.62$\pm$ 0.13 & 79.62 $\pm$ 0.13  & \textbf{85.90 $\pm$ 0.10}\\
Places365 & 90.42 $\pm$ 0.09  &   90.37 $\pm$ 0.0 &  \textbf{99.98 $\pm$ 0.01}
  & 98.10 $\pm$ 0.02  & 98.09 $\pm$ 0.02 & \textbf{99.99 $\pm$ 0.00}\\
\hline
\end{tabular}
\end{adjustbox}
\end{center}
\end{table*}

\subsubsection{C.1.4. Model details for state-of-the-art results}
\label{apdx:sota_models}

\textbf{\\ Model details for CIFAR-10. } We use the same WRN architecture as in the other self-supervised and unsupervised OOD detectors\footnote{We use the wrn architecure from \url{https://github.com/hendrycks/ss-ood/blob/master/multiclass_ood/models/wrn.py}}. 
The architecture of the encoder and the decoder blocks are the same as in the AVT model for ablation studies.
Again, the encoder block is inserted after the last block of the WRN, followed by the global average pooling layer. 
The original and transformed  images $x$, $g(x), g \in G$  are fed in this architecture. The average-pooled features of the original and the transformed images are concatenated and fed into the decoder; a two-layer fully connected network. We use ReLU activation after the first layer, and the second layer is a softmax layer predicting the class of the transformation. 
Similar to~\citet{GOAD}, we use cross-entropy loss with center-triplet loss~\citep{center_triplet_loss} to train the model.~\citet{GOAD} propose that center-triplet loss (used along with cross-entropy loss to stabilize the training) yields better performance than cross-entropy loss. We also found the same in our experiments.

\textbf{Model details and results on CIFAR-100. } Here also, we use the same architecture of the AVT model, i.e. ResNet-18 that has been used for one-class OOD detection on CIFAR-100 by the other SOTA unsupervised and self-supervised detectors. All settings except for the model architecture (ResNet-18 instead of WRN) are same as the one-class OOD detection experiments on CIFAR-10.
Table~\ref{cifar_100_sota-results} shows the comparison of AUROC results for one-class OOD detection results on individual classes of CIFAR-100. We could achieve competitive results on most of the classes and the overall mean.

\begin{table}[!t]

\caption {\label{cifar_100_sota-results}\footnotesize{AUROC for one-class OOD detection on 20 superclasses of CIFAR-100  by SOTA, ICAD($|\mathcal V(x)|=1$) and Ours($|\mathcal V(x)|=5$).}}

\setlength{\tabcolsep}{3pt}
\begin{adjustbox}{width=1\columnwidth,right}
\begin{tabular}{cccccccccc}
\toprule
Class &  SVM  & GM & AUX & AUX+Trans & GOAD & CSI & ICAD & Ours \\
\hline
0  &  68.4 &  74.7 &  78.6 & 79.6 & 73.9 & \textbf{86.3} & 80.56$\pm$ 0.16 & \underline{81.09 $\pm$ 0.15}   \\
1  &  63.6 & 68.5 & 73.4 & 73.3 & 69.2 & \textbf{84.8} & 77.28 $\pm$ 0.14  & \underline{77.79 $\pm$ 0.13}   \\
2  &  52.0 & 74.0 & 70.1 & 71.3 & 67.6 & \textbf{88.9} & 75.72 $\pm$ 0.08 &  \underline{75.90 $\pm$ 0.07}   \\
3  &  64.7 & \underline{81.0} & 68.6 & 73.9 & 71.8 & \textbf{85.7} & 65.06$\pm$ 0.25 &   65.40 $\pm$ 0.13   \\
4  &  58.2 & 78.4  & 78.7 & 79.7 & 72.7 & \textbf{93.7} & 81.86 $\pm$ 0.21 & \underline{82.16 $\pm$ 0.11}   \\
5  &  54.9 & 59.1  & 69.7 & \underline{72.6} & 67.0 & \textbf{81.9} & 67.79 $\pm$ 0.21 & 68.15 $\pm$ 0.13   \\
6  &  57.2 & 81.8  & 78.8 & 85.1 & 80.0 & \textbf{91.8} & 84.96  $\pm$ 0.21 & \underline{85.47 $\pm$ 0.09}  \\
7  &  62.9 & 65.0  & 62.5 & \underline{66.8} & 59.1 & \textbf{83.9} & 64.35$\pm$ 0.44 & 64.70 $\pm$ 0.15   \\
8  &  65.6 & 85.5  & 84.2 & \underline{86.0} & 79.5 & \textbf{91.6} & 84.05$\pm$ 0.13 & 84.60 $\pm$ 0.08   \\
9  &  74.1 & 90.6  & 86.3 & 87.3 & 83.7 & \textbf{95.0} & 94.43$\pm$ 0.09 & \underline{94.70 $\pm$ 0.05}   \\
10  & 84.1 & 87.6  & 87.1 & 88.6 & 84.0 & \textbf{94.0} & 88.80$\pm$ 0.07 & \underline{88.92 $\pm$ 0.04}   \\
11  & 58.0 & \underline{83.9}  & 76.2 & 77.1 & 68.7 & \textbf{90.1} & 78.98 $\pm$ 0.20 &79.44 $\pm$ 0.13   \\
12  & 68.5 & 83.2  & 83.3 & 84.6 & 75.1 & \textbf{90.3} & 84.76 $\pm$ 0.10 & \underline{85.26 $\pm$ 0.07}   \\
13  & \underline{64.6} & 58.0  & 60.7 & 62.1 & 56.6 & \textbf{81.5} & 59.91 $\pm$ 0.23 & 59.94 $\pm$ 0.11   \\
14  & 51.2 & 92.1  & 87.1 & 88.0 & 83.8 & \textbf{94.4} & 92.71 $\pm$ 0.14 & \underline{93.08 $\pm$ 0.10}   \\
15  & 62.8 & 68.3  & 69.0 & \underline{71.9} & 66.9 & \textbf{85.6} & 67.57$\pm$ 0.19 & 67.86 $\pm$ 0.17   \\
16  & 66.6 & 73.5  & 71.7 & \underline{75.6} & 67.5 & \textbf{83.0} & 72.75$\pm$ 0.31 & 73.19$\pm$ 0.15    \\
17  & 73.7 & 93.8  & 92.2 & 93.5 & 91.6 & 97.5 & \underline{98.10 $\pm$ 0.09} & \textbf{98.28 $\pm$ 0.03}   \\
18  & 52.8 & 90.7  & 90.4 & 91.5 & 88.0 & \textbf{95.9} & 93.28 $\pm$ 0.11 & \underline{93.66 $\pm$ 0.08}   \\
19  & 58.4 & 85.0  & 86.5 & 88.1 & 82.6 & \textbf{95.2} & 93.36 $\pm$ 0.10 & \underline{93.67$\pm$ 0.04} \\
\midrule
Mean & 63.1 & 78.7 & 77.7 & 79.8 & 74.5 & \textbf{89.6} & 80.31 &   \underline{80.66} \\
\bottomrule
\end{tabular}
\end{adjustbox}

\end{table}

\subsubsection{C.1.5. Details of~\citet{baseline}'s SOTA OOD detection method (SBP) for OOD detection on CIFAR-10 as iD.}
\label{apdx:sota_sbp}
SBP uses the maximum softmax score from a classifier trained on the iD data for OOD detection. As proposed by~\citet{qi2019}, we freeze the AVT encoder trained on CIFAR-10 from ablation studies and train a classifier on top of it. The architecture of the classifier is same as the decoder's architecture in the AVT model for CIFAR-10. The trained classifier achieved test accuracy of 91.46\% on CIFAR-10 and was used in detection. 

\subsubsection{C.1.6. Generalizability of iDECODe with respect to the transformation set $G$. }
\label{apdx:gen_trans_set}
Table~\ref{table:gen_trans_set} shows generalizability of iDECODe with respect to the set $G$ of transformations. Here we compare the performance of iDECODe (on CIFAR-10 as iD) with two sets of transformations and the corresponding base NCM. The first set of transformations is the set of projective transformations (PT) from the ablation studies with mean square error as the base NCM. The second set of transformations is the set of four rotation ranges (RS) from SOTA experiments with cross-entropy loss as the base NCM.

\textbf{Discussion on results. } The results show that SVHN and Places365 are better detected with $G=PT$, and others (LSUN, Imagenet, and CIFAR-100) are better detected with $G=RS$. We believe that it is an important observation in the domain of self-supervised OOD detection that some OOD sets are better detected with one transformation set vs the other. A possible reason for these results could be the diversity in the OOD-ness of different datasets (ex. CIFAR-100 vs Places365) leading to the difference in detection abilities on these datasets~\cite{kaur2021all}. Further investigation on this will be our future work.

\begin{table}[]
\begin{center}
\setlength{\tabcolsep}{5pt}
\caption{\footnotesize{Comparing AUROC of iDECODe for $G$ = set of projective transformations (PT) and $G$ = set of rotations (RT).}}

\label{table:gen_trans_set}
\begin{adjustbox}{width=1\columnwidth}
\begin{tabular}{c|cc|cc}
\hline
$D_{out}$   &   \multicolumn{2}{c|}{$|\mathcal V(x)|=1$}  & \multicolumn{2}{c}{$|\mathcal V(x)|=5$}     \\ 
\cline{2-3} \cline{4-5}
      & $G=PT$  & $G=RS$                                                             & $G=PT$  & $G=RS$                                          \\ 
 \cline{2-3} \cline{4-5}
SVHN  &  92.86 $\pm$ 0.16  &  92.09 $\pm$  0.02  &95.70 $\pm$ 0.07   &  92.37 $\pm$ 0.02  \\ 
LSUN &  79.47$\pm$  0.26 &   86.65 $\pm$  0.06 &  85.98 $\pm$ 0.13   &  87.52 $\pm$ 0.05  \\
   
ImageNet &  82.21 $\pm$  0.18  &  88.59 $\pm$  0.05 &  87.97 $\pm$ 0.15  &  89.43 $\pm$ 0.05 \\

CIFAR100 & 72.60 $\pm$  0.30  & 82.46 $\pm$  0.04 & 78.04 $\pm$ 0.20   &  83.08 $\pm$ 0.04  \\

Places365 &  96.87 $\pm$  0.06  & 86.73 $\pm$  0.04 &  99.98 $\pm$ 0.01   &  88.15 $\pm$ 0.03  \\
\hline
\end{tabular}
\end{adjustbox}
\end{center}

\end{table}

\subsection{C.2. Audio}
\label{apdx:audio}
As shown in Table~\ref{tab:audio_sets}, based on the similarity among the classes, we group twenty classes of the FSDNoisy18k audio dataset into four sets. This grouping is performed based on the similarity of the classes, e.x. acoustic\_guitar and bass\_guitar are in the set 0 and
fart, fire and fireworks are in the set 1. 

We use the same VGG classifier trained with data augmentation (for iDECODe\footnote{Our code is build on top of~\citet{audio_ood}'s code for training VGG classifier on the FSD audio dataset from \url{https://github.com/tqbl/ood_audio}.}) on the iD set for obtaining the SBP results.

\begin{table}[]
\caption{\footnotesize{Grouping of FSDNoisy18k classes into four sets.}} 

\begin{adjustbox}{width=1\columnwidth,center}
\begin{tabular}{|c|c|}
\hline
$D_{in}$  & \multicolumn{1}{c|}{FSDNoisy18k classes}  \\ 

\hline
Set 0 &  Acoustic\_guitar, Bass\_guitar, Clapping, Coin\_(dropping), Crash\_cymbal\\

Set 1 & Dishes\_and\_pots\_and\_pans, Engine, Fart, Fire, Fireworks \\
Set 2 & Glass, Hi-hat, Piano, Rain, Slam\\

Set 3 & Squeak, Tearing, Walk\_or\_footsteps, Wind,
    Writing \\
\hline
\end{tabular}
\end{adjustbox}
\label{tab:audio_sets} 

\end{table}

\subsection{C.3. Adversarial}
\label{apdx:adv}
Here we describe the details of SOTA methods used for comparison with iDECODe on adversarial detection. 

\textbf{\\ Supervised detectors.} We compare with KD+PU, LID and Mahala as the supervised adversarial detectors. Kernel Density (KD)~\citep{kd_pu} detects datapoints lying in the low Guassian density regions as adversarial. Specifically, KD uses the following score to detect an input $x$ as adversarial:
\begin{equation*}
    KD(x) = \frac{1}{|X_t|}\sum_{x_i\in X_t} k(x_i,x).
\end{equation*}
Here $t$ is the predicted class label of $x$, $X_t$ is the set of training samples with the label $t$, and $k_{\sigma}(x,y) \sim \exp({-||x-y||^2/\sigma^2})$.  Predictive Uncertainty (PU) is the maximum value of the softmax score predicted by the classifier. KD+PU is a logistic regression detector trained on the scores from KD and PU to detect adversarial inputs. Local Intrinsic Dimensionality (LID)~\citep{lid} uses distance distribution of the input's neighboring samples from the training data to detect adversaries. Specifically, LID computes the following score for an input $x$ to detect it as an adversarial input:
\begin{equation*}
    \widehat{LID}(x) = -\left(\frac{1}{k}\sum_{i=1}^{k}\log\frac{r_i(x)}{r_k(x)}\right)^{-1}.
\end{equation*}
Here $r_i(x)$ is the distance between $x$ and $i^{th}$ training sample from the set of $k$ nearest neighbors and $r_k(x)$ is the distance from farthest sample in the set. Logistic regression detector trained on LID scores from all the layers is used to detect adversarial inputs.
Mahala~\citep{mahalanobis} computes Mahalanobis distance of the noisy input in the feature space of the training data as a score to detect adversarial inputs. Logistic regression detector trained on mahalanobis distance from all the layers is used to detect adversarial inputs.

\textbf{\\ 
Unsupervised detectors.} We use Odds and AE as the unsupervised adversarial detectors. Odds~\citep{odds_testing} propose statistical testing for adversarial detection based on the insight that feature variation caused by adding noise to the input tends to have a characteristic direction for adversarial samples. Specifically, they measure the change in pair-wise class logits caused by adding noise to the input to detect adversaries. AE~\citep{ae_layers} makes use of autoencoders trained on the feature space from every layer of the classifier to detect adversarial inputs. Specifically, they use reconstruction error and latent norm from the trained autoencoders as scores for adversarial detection.

\subsection{C.4. Computational resources}
\label{apdx:compute}
We run all our experiments on a server with CUDA version 11.1 and four GPU cards. The specifications of  all the four GPU cards are same and are as follows:
\begin{enumerate}
    \item Card: NVIDIA Corporation [GeForce RTX 2080 Ti Rev. A].
    \item Memory: 11019MiB.
\end{enumerate}

\section{D. Details on measurability}
\label{apdx:meas_det}

Here we provide some standard details on the measurability issues from our paper. 
We assume we have a probability space $(\Omega, B, P)$, where $\Omega$ is a base space, $B$ is a sigma-algebra on $\Omega$, and  $P:B \to [0,1]$ is a probability measure on $B$.
The set $G$ of transformations is a subset of all maps $\mathcal{X} \to \mathcal{X}$, and is also equipped with a sigma-algebra $B_G$ of measurable subsets of $G$. We assume that $\mathcal{X}$, the feature space, is  equipped with a sigma-algebra $B_X$, and that each $g\in G$ is measurable with respect to $B_X$. 

Choosing a random $g\in G$ is accomplished by a measurable function $g: X \times \Omega \to G$ (measurable with respect to the product sigma-algebra $B_X\times B$ over the input and $B_G$ over the output), which assigns to each $\omega \in \Omega$  a specific $g=g( \cdot; \omega) \in G$. The distribution of $g(\cdot, \omega)$ when $\omega \sim P$ is denoted as $Q_G:B_G \to [0,1]$, and is a probability measure on $(G,B_G)$. Then, we can construct the product measure $P^{(n+1),(k+1)}$, and thus also the product measure $Q_G^{(n+1),(k+1)}$, allowing us to sample iid $g_{ji}, j\in \{m+1, \ldots, l, x\}, i=1, \ldots, n$. Moreover, we assume that the map $g\to g'$ mapping the input transform to the output transform is measurable over the sigma algebra $B_G$ for the input and output. 

We assume that $M:\mathcal{X}\to \mathcal{X}$ is measurable with respect to $B_X$, and that the loss function $L: \mathcal{X} \times \mathcal{X} \to \mathbb{R}$ is measurable as a function between the sigma-algebra $B_X \times B_X$ on $\mathcal{X} \times \mathcal{X}$ and the Borel-sigma algebra on $\mathbb{R}$.

Finally, we also assume that $A(X_\mathrm{tr}, \cdot, \cdot): \mathcal{X} \times G\to \mathbb{R}$ is measurable with respect to the sigma-algebra $B_X\times B_G$, and that the aggregation function $F$ is measurable with respect to the Borel sigma-algebras for its input and the output.

\end{document}